\documentclass[conference]{IEEEtran}
\IEEEoverridecommandlockouts
\usepackage{cite}
\usepackage{amsmath,amssymb,amsfonts}
\usepackage{algorithmic}
\usepackage{graphicx}
\usepackage{textcomp}
\usepackage{xcolor}
\usepackage{subcaption}
\usepackage{graphicx}
\usepackage{subcaption}
\usepackage{float}
\usepackage{placeins}
\usepackage{dblfloatfix}
\def\BibTeX{{\rm B\kern-.05em{\sc i\kern-.025em b}\kern-.08em
    T\kern-.1667em\lower.7ex\hbox{E}\kern-.125emX}}
\begin{document}

\title{Geometry-Aware Predictive Control for Walk-to-Run Bipedal Obstacle Crossing

}
\author{
\IEEEauthorblockN{1\textsuperscript{st} Aashish Sahu}
\IEEEauthorblockA{
\textit{Department of Mechanical and Aerospace Engineering} \\
\textit{Indian Institute of Technology Hyderabad} \\
Hyderabad, India \\
me22resch11011@iith.ac.in}
\and
\IEEEauthorblockN{2\textsuperscript{nd} R. Prasanth Kumar}
\IEEEauthorblockA{
\textit{Department of Mechanical and Aerospace Engineering} \\
\textit{Indian Institute of Technology Hyderabad} \\
Hyderabad, India \\
rpkumar@mae.iith.ac.in}
}

\maketitle

\begin{abstract}
This paper presents a geometry-aware predictive control framework for bipedal obstacle crossing across walking and running regimes. The proposed approach integrates a phase-structured trajectory generator with a constrained model predictive control (MPC) tracking layer to generate coordinated pelvis and swing-foot motions during obstacle traversal. The crossing process is represented through five interpretable phases: pre-crossing, leading-leg clearance, straddle, trailing-leg clearance, and recovery. Within this structure, swing-foot elevation is adapted as a function of obstacle height, while lateral stance widening and pelvis repositioning are introduced as functions of obstacle depth to improve transverse clearance and body alignment. A continuous gait-parameter evolution enables smooth walk-to-run transition without discrete controller switching. The generated references are tracked using MPC to regulate motion smoothness, clearance feasibility, touchdown behavior, and lateral admissibility. The framework is evaluated in reduced-order simulation over five progressively difficult obstacles under walking and running conditions. The robot successfully crosses the first four obstacles, achieving minimum body--obstacle clearances of 0.036, 0.036, 0.059, and 0.017~m, respectively. The fifth and most demanding obstacle leads to collision, identifying the current operating boundary of the proposed reduced-order formulation. Overall, the method provides a compact, interpretable, and computationally efficient basis for intelligent obstacle-aware bipedal locomotion.
\end{abstract}

\begin{IEEEkeywords}
Bipedal locomotion, obstacle crossing, model predictive control, geometry-aware trajectory generation, walk-to-run transition,legged robotics.
\end{IEEEkeywords}

\section{Introduction}

Bipedal locomotion enables robotic systems to operate in human-centered environments containing obstacles, uneven terrain, and discontinuous footholds. Compared with wheeled or tracked platforms, bipedal robots can negotiate certain obstacles by stepping over them while maintaining forward progression. However, successful obstacle crossing requires coordinated swing-leg clearance, pelvis motion, foot placement, body posture, and contact timing. These requirements become more demanding during walking-to-running transitions, as gait frequency, step length, duty factor, and available support time change with locomotion speed. Raising the swing foot alone is therefore insufficient: the leading and trailing legs must clear the obstacle while the pelvis and lateral foot placements remain compatible with the crossing sequence. This motivates a control-oriented formulation that connects obstacle geometry to coordinated motion references across different gait regimes.

Classical locomotion methods employ reduced-order models and balance-related formulations, including zero-moment-point preview control, capture-point regulation, spring-mass running templates, and hybrid or human-inspired walking control~\cite{kajita2003,pratt2006,raibert1986,blickhan1989,westervelt2007,ames2014}. These approaches provide established foundations for gait generation, stabilization, and recovery. Optimization-based locomotion planning, adaptive foot-placement MPC, differential dynamic programming, and contact-aware trajectory optimization further support motion generation under dynamic and environmental constraints~\cite{kuindersma2016,diedam2008,herdt2010, sahu2024hexacopter, tassa2014,posa2014,mordatch2012}. In particular, predictive adaptation of foot placement is already well established; consequently, the use of MPC alone does not constitute the contribution of this work. Instead, the present study examines how obstacle-relative geometric references can be constructed and coordinated before predictive tracking, while retaining an interpretable representation of the crossing motion.

Human locomotion studies show that obstacle negotiation involves anticipatory gaze, stepping adjustments, leading- and trailing-leg clearance, and changes in body posture~\cite{patla1997,chen1991,mcfadyen1991}. These observations motivate a phase-structured representation in which obstacle geometry influences more than the vertical swing trajectory. Footstep planning and swing-height adaptation address important aspects of traversal, but the present formulation additionally coordinates lateral stance adjustment and pelvis repositioning with the progression of both legs. The specific research problem is therefore the construction of a compact reference-generation mechanism that jointly incorporates obstacle height, transverse extent, crossing phase, and continuous gait evolution. \cite{sahu2024armleg, sahu2026integrated, sahu2026dual}This scope distinguishes the proposed integration from its established control components without implying that existing locomotion controllers are inherently unable to negotiate obstacles.

To address this problem, a geometry-aware trajectory generator is coupled with a constrained model predictive control tracking layer. Obstacle traversal is organized into five phases: pre-crossing, leading-leg clearance, straddle, trailing-leg clearance, and recovery. Swing-foot elevation is parameterized by obstacle height, while stance widening and pelvis offset depend on the obstacle's transverse extent, termed depth in the adopted geometry definition. Smooth interpolation connects the swing and hold intervals, and continuous variation of gait parameters links walking and running references without discrete controller switching. Separating reference construction from tracking allows the geometric coordination to remain explicit, while MPC uses preview and penalties on tracking error, control effort, and input variation to regulate the resulting motion. The procedure is selected for its interpretable coordination structure rather than a demonstrated computational advantage over alternative controllers.

The principal contribution is the integration of geometry-dependent swing clearance, lateral stance adjustment, pelvis repositioning, and continuous gait evolution within a common MPC-tracked crossing sequence. The framework is evaluated in reduced-order MATLAB simulation using five progressively taller obstacles encountered in walking and running regimes. The first four obstacles are crossed with positive minimum sampled body--obstacle clearance, whereas the fifth produces collision before the intended elevated posture is attained. Retaining this failed encounter documents a limitation of the tested configuration rather than implying unrestricted traversal capability. Because obstacle depth is fixed and matched baseline or ablation results are not available, the reported evaluation establishes nominal feasibility but does not isolate the benefit of each component or demonstrate comparative superiority. The successful and failed cases provide a basis for subsequent controlled comparisons, failure diagnostics, and extension to full-body contact dynamics.

\section{Proposed Methodology}

\begin{figure}[t]
    \centering
    \includegraphics[width=0.98\linewidth]{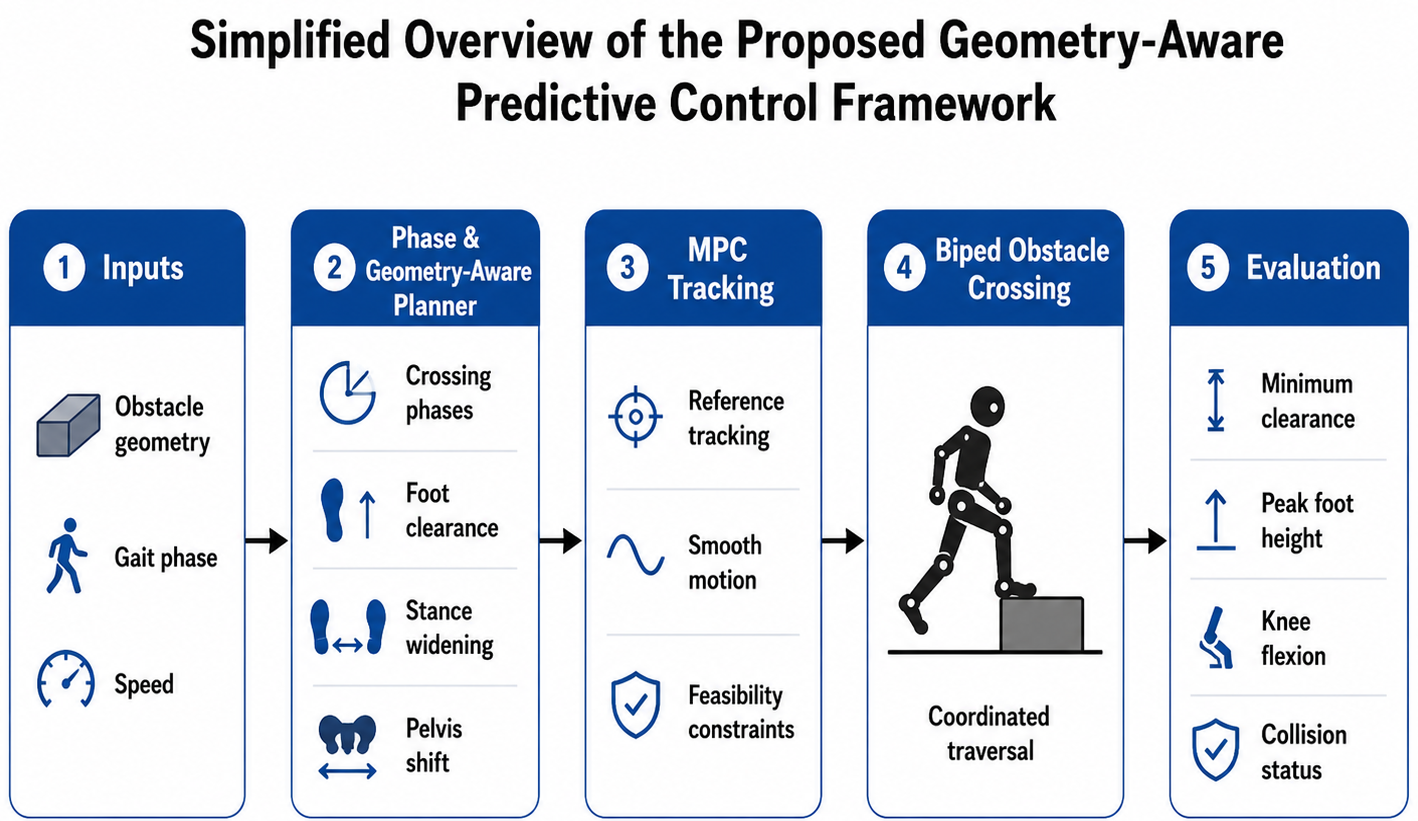}
    \caption{Geometry-aware reference generation and constrained
    predictive tracking for phase-structured obstacle crossing.}
    \label{fig:framework}
\end{figure}

The proposed framework couples a geometry-aware trajectory generator with a constrained model predictive control (MPC) tracking layer, as illustrated in Fig.~\ref{fig:framework}. The trajectory generator coordinates obstacle-relative pelvis and foot references, while MPC regulates tracking and command variation within prescribed bounds. The methodological contribution is this phase-dependent coordination of geometric adaptations, rather than a new MPC algorithm.

\subsection{Obstacle Geometry and Crossing Phase}

The reduced-order biped is described by pelvis position
\(\mathbf{p}_p=[x_p,y_p,z_p]^T\) and foot positions
\(\mathbf{p}_{L,R}=[x_{L,R},y_{L,R},z_{L,R}]^T\), where
\(x\), \(y\), and \(z\) denote forward, lateral, and vertical
directions. Each obstacle is represented as
\begin{equation}
\mathcal{O}_i=
\left\{\mathbf{p}\in\mathbb{R}^3:
\begin{array}{l}
x_i\leq x\leq x_i+w_i,\\
|y|\leq d_i/2,\quad 0\leq z\leq h_i
\end{array}
\right\},
\label{eq:obstacle}
\end{equation}
where \(h_i\) is height, \(w_i\) is extent along travel, and
\(d_i\) is transverse extent, termed depth in the adopted
geometry definition. Thus, the lateral adaptation law depends
on transverse extent rather than forward obstacle thickness.

Crossing progression is parameterized by
\begin{equation}
s=\frac{x_p-x_{\mathrm{start},i}}
{x_{\mathrm{end},i}-x_{\mathrm{start},i}},
\label{eq:crossing_phase}
\end{equation}
with \(x_{\mathrm{end},i}>x_{\mathrm{start},i}\). The crossing
sequence comprises pre-crossing, leading-leg clearance,
straddle, trailing-leg clearance, and recovery. This
decomposition coordinates both legs with pelvis progression
instead of treating clearance as an isolated foot-lifting task.

Walking and running references are connected through continuous
variation of
\(\boldsymbol{\eta}(v)=[f_g(v),L_s(v),\beta(v),h_f(v)]^T\),
where the components denote gait frequency, step length, duty
factor, and nominal swing height. For time-varying frequency,
a continuous accumulated gait phase is defined by
\begin{equation}
\phi(t)=\mathrm{mod}
\left(\phi_0+\int_0^t f_g(v(\tau))\,\mathrm{d}\tau,\,1\right),
\label{eq:gait_phase}
\end{equation}
which reduces to \(\mathrm{mod}(\phi_0+f_gt,1)\) at constant
frequency. Continuous gait parameterization avoids discrete
reference switching; dynamic feasibility remains dependent on
the model and imposed constraints.

\subsection{Geometry-Aware Motion References}

Each swing foot follows a hold--swing--hold profile, with
leg-specific swing boundaries \(a<b\). Within \(a<s<b\),
define \(u_s=(s-a)/(b-a)\) and
\(\hat{u}_s=3u_s^2-2u_s^3\). The forward reference is
\begin{equation}
x_f(s)=
\begin{cases}
x_{\mathrm{st}}, & s\leq a,\\
(1-\hat{u}_s)x_{\mathrm{st}}+\hat{u}_s x_{\mathrm{ld}},
& a<s<b,\\
x_{\mathrm{ld}}, & s\geq b,
\end{cases}
\label{eq:forward_reference}
\end{equation}
where \(x_{\mathrm{st}}\) and \(x_{\mathrm{ld}}\) are the
departure and landing positions. The cubic interpolation has
zero endpoint slope, providing a smooth forward transition
between the hold intervals. The vertical reference is
\begin{equation}
z_f(s)=
\begin{cases}
z_{\mathrm{clr},i}\sin^\gamma(\pi\hat{u}_s), & a<s<b,\\
0, & \mathrm{otherwise},
\end{cases}
\label{eq:vertical_reference}
\end{equation}
with commanded apex
\begin{equation}
z_{\mathrm{clr},i}=h_i+\Delta h+k_hh_i.
\label{eq:height_adaptation}
\end{equation}
Here, \(\Delta h\) provides a fixed vertical margin, \(k_h\)
scales the margin with obstacle height, and \(\gamma\) shapes
the swing profile. This law specifies the desired foot apex;
it does not independently guarantee whole-body clearance.

The desired half-stance width is
\begin{equation}
y_i^\star=
\min\left(y_{\max},
\max\left(y_0,\frac{d_i}{2}+\Delta y\right)\right),
\label{eq:stance_target}
\end{equation}
where \(y_0\) is nominal half-stance width, \(\Delta y\) is a
lateral margin, and \(y_{\max}\) limits widening. A bounded
crossing envelope \(B(s)\in[0,1]\) blends nominal and
obstacle-adapted placement:
\begin{equation}
\begin{aligned}
y_w(s)&=(1-B(s))y_0+B(s)y_i^\star,\\
y_L^{\mathrm{ref}}&=+y_w(s),\qquad
y_R^{\mathrm{ref}}=-y_w(s).
\end{aligned}
\label{eq:lateral_references}
\end{equation}
The same envelope coordinates the obstacle-induced pelvis
offset:
\begin{equation}
\begin{aligned}
y_p^{\mathrm{ref}}(t)
&=A_y\sin(2\pi\phi(t))+\sigma_i\delta_{p,i}B(s),\\
\delta_{p,i}&=\min(\delta_{p,\max},k_pd_i),
\end{aligned}
\label{eq:pelvis_reference}
\end{equation}
where \(A_y\) is nominal lateral oscillation amplitude,
\(\sigma_i\in\{-1,+1\}\) selects the offset direction, and
\(k_p\) and \(\delta_{p,\max}\) determine its scaling and limit.
These references coordinate lateral placement and pelvis
motion; their individual performance benefits require
controlled ablation tests.

\subsection{Predictive Tracking and Clearance Assessment}

The generated references are tracked using
\begin{equation}
\mathbf{x}_{k+1}=A\mathbf{x}_k+B\mathbf{u}_k,
\label{eq:prediction_model}
\end{equation}
where \(\mathbf{x}_k\) contains pelvis and swing-foot positions
and velocities, and \(\mathbf{u}_k\) contains acceleration-like
inputs. At update \(k\), the horizon objective is
\begin{equation}
J_k=\sum_{\ell=0}^{N-1}
\left[
\|\mathbf{e}_{\ell|k}\|_Q^2+
\|\mathbf{u}_{\ell|k}\|_R^2+
\|\Delta\mathbf{u}_{\ell|k}\|_{R_\Delta}^2
\right],
\label{eq:mpc_objective}
\end{equation}
where
\(\mathbf{e}_{\ell|k}=
\mathbf{x}_{\ell|k}-\mathbf{x}_{\ell|k}^{\mathrm{ref}}\),
\(\|\mathbf{q}\|_W^2=\mathbf{q}^TW\mathbf{q}\), and
\(\Delta\mathbf{u}_{\ell|k}=
\mathbf{u}_{\ell|k}-\mathbf{u}_{\ell-1|k}\).
For \(\ell=0\), the previous input is the last applied command.
The weights penalize tracking error, control effort, and
command variation, respectively.

Optimization is subject to the prediction model, state/input
bounds, and phase-dependent restrictions on swing clearance,
touchdown velocity, lateral placement, and pelvis motion.
The first optimized input is applied before the horizon is
advanced. MPC is selected to incorporate preview and these
tracking restrictions within one optimization; it does not
provide a full-body collision or dynamic-stability guarantee
for the reduced-order representation.

Traversal is assessed using minimum signed clearance,
\begin{equation}
d_{\min}(t)=
\min_{i,j}\mathrm{sdist}
\bigl(\mathbf{p}_j(t),\mathcal{O}_i\bigr),
\label{eq:minimum_clearance}
\end{equation}
where \(\mathbf{p}_j(t)\) denotes an evaluated body point.
Positive values indicate separation of the evaluated points,
zero denotes contact, and negative values indicate penetration.
This sampled metric characterizes the reported simulation
outcomes rather than guaranteeing continuous swept-volume
collision avoidance.

\section{Simulation Setup}

\begin{figure*}[!t]
    \centering
    \begin{subfigure}[t]{0.19\textwidth}
        \centering
        \includegraphics[width=\linewidth]{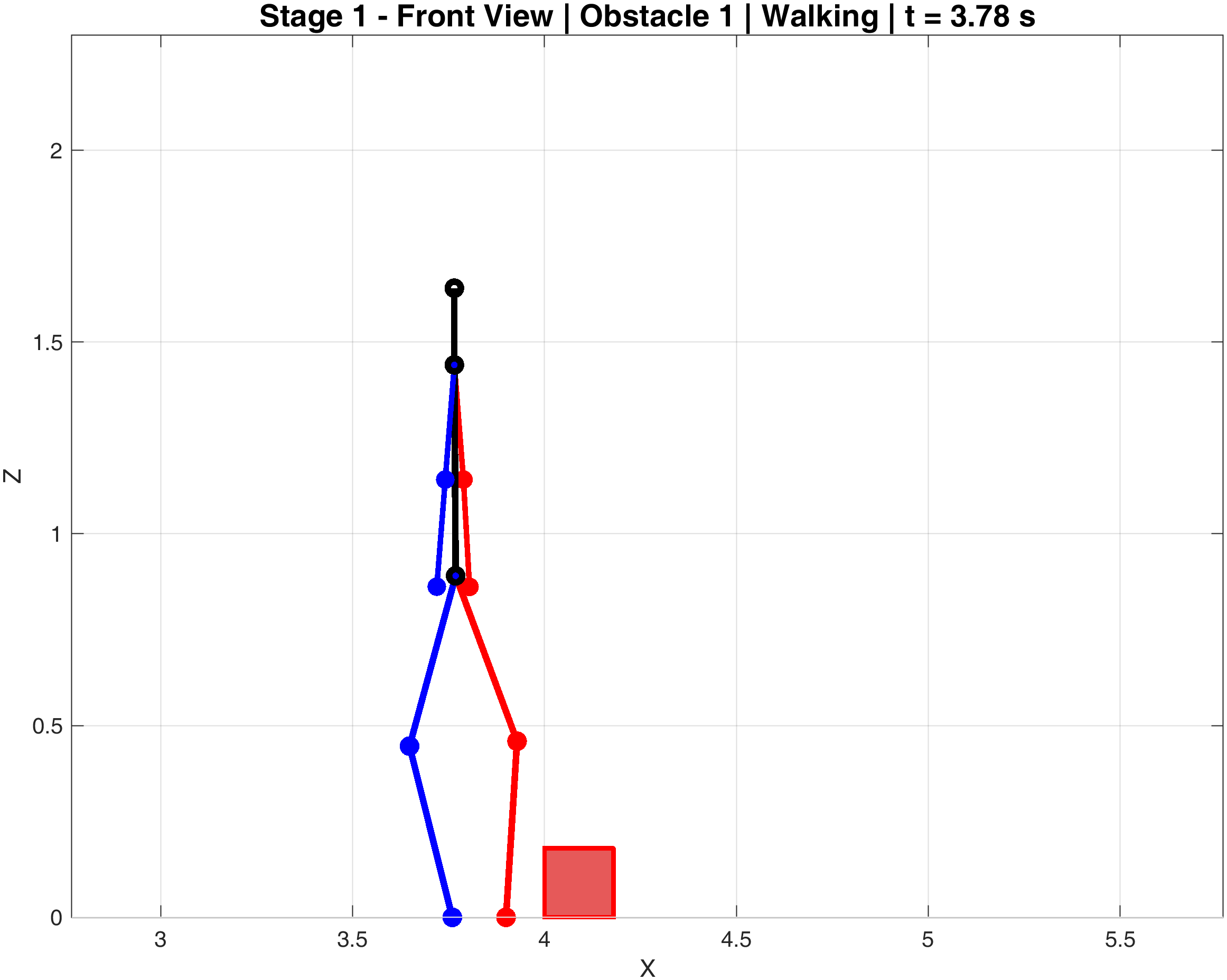}
        \caption{Stage 1}
    \end{subfigure}
    \hfill
    \begin{subfigure}[t]{0.19\textwidth}
        \centering
        \includegraphics[width=\linewidth]{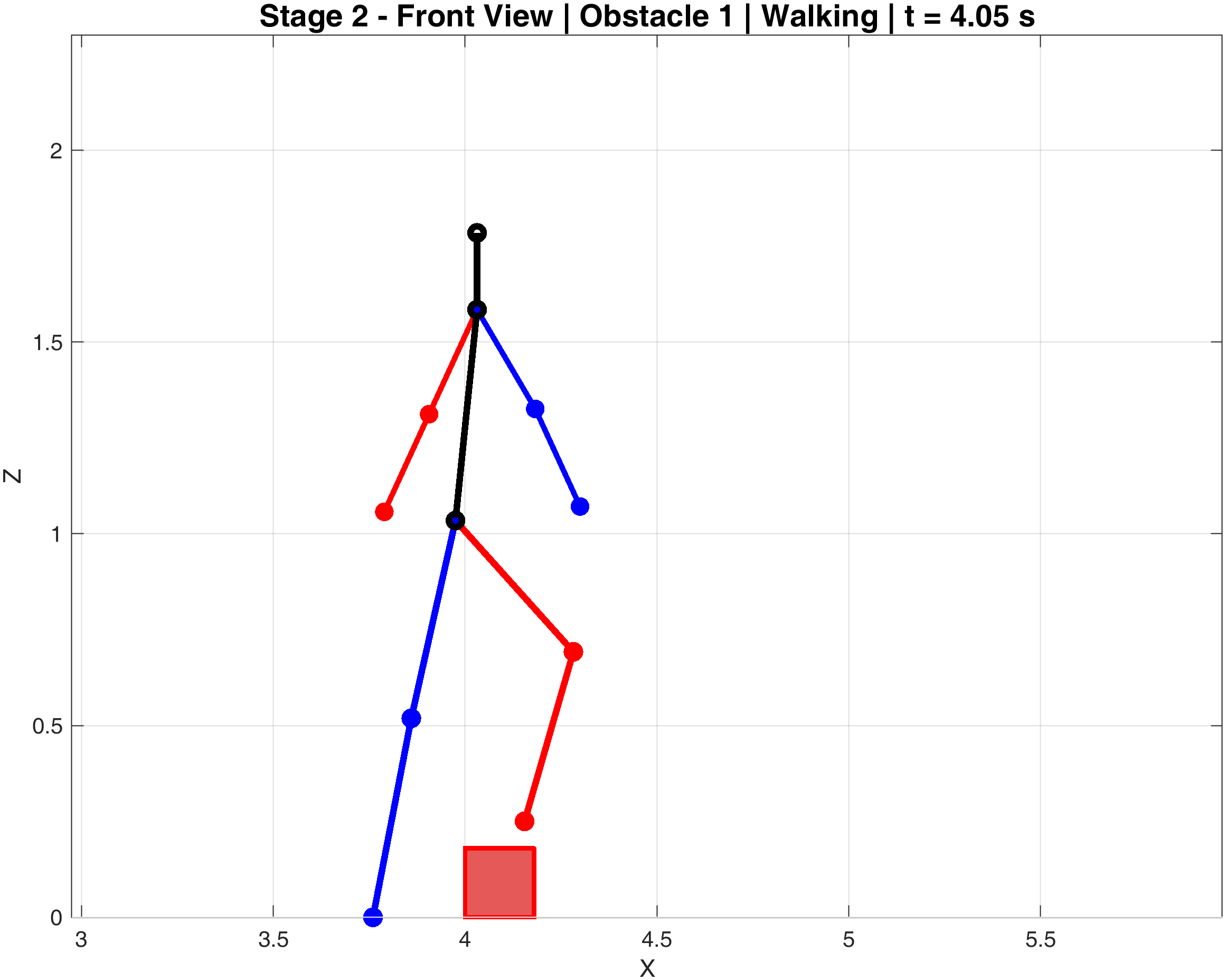}
        \caption{Stage 2}
    \end{subfigure}
    \hfill
    \begin{subfigure}[t]{0.19\textwidth}
        \centering
        \includegraphics[width=\linewidth]{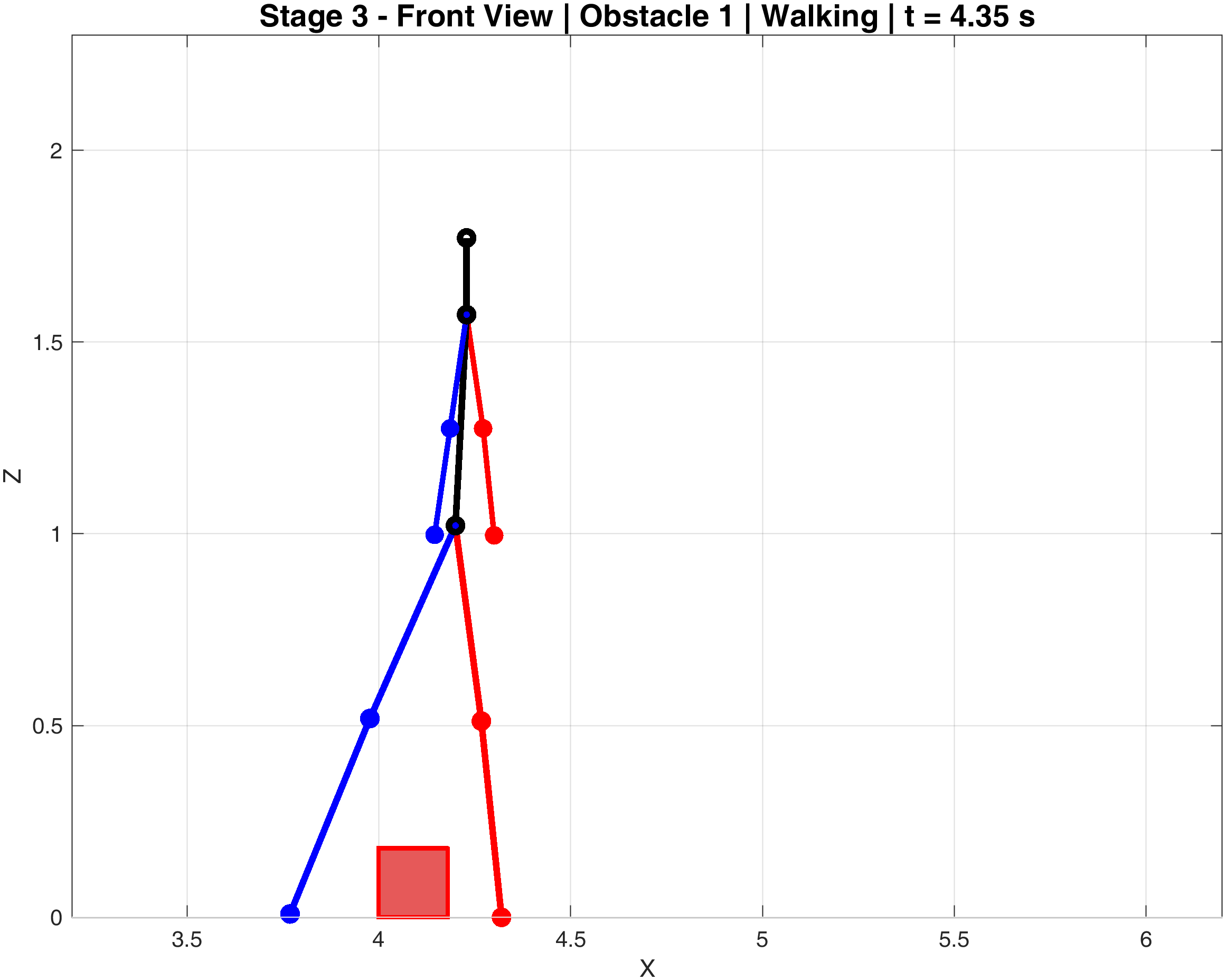}
        \caption{Stage 3}
    \end{subfigure}
    \hfill
    \begin{subfigure}[t]{0.19\textwidth}
        \centering
        \includegraphics[width=\linewidth]{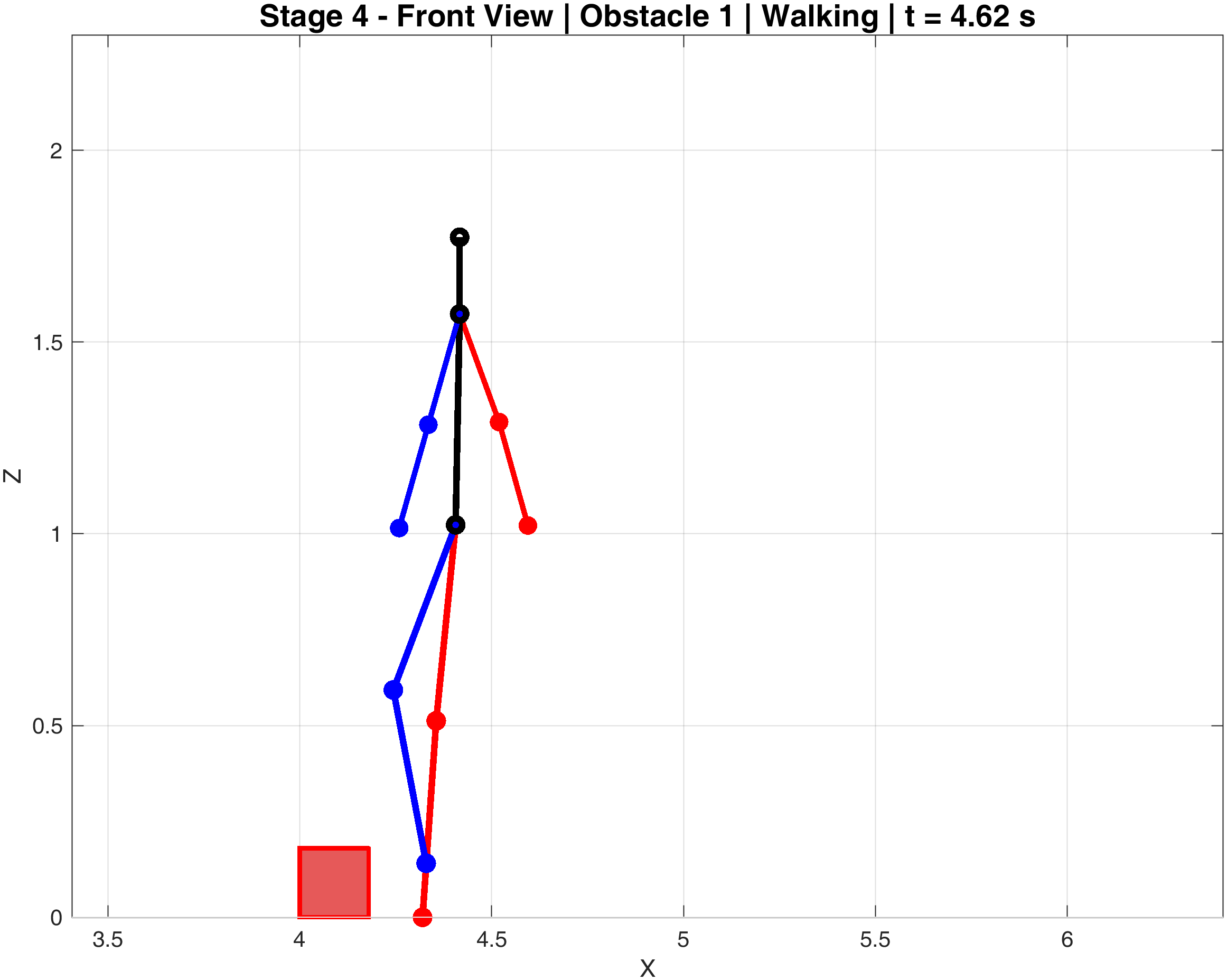}
        \caption{Stage 4}
    \end{subfigure}
    \hfill
    \begin{subfigure}[t]{0.19\textwidth}
        \centering
        \includegraphics[width=\linewidth]{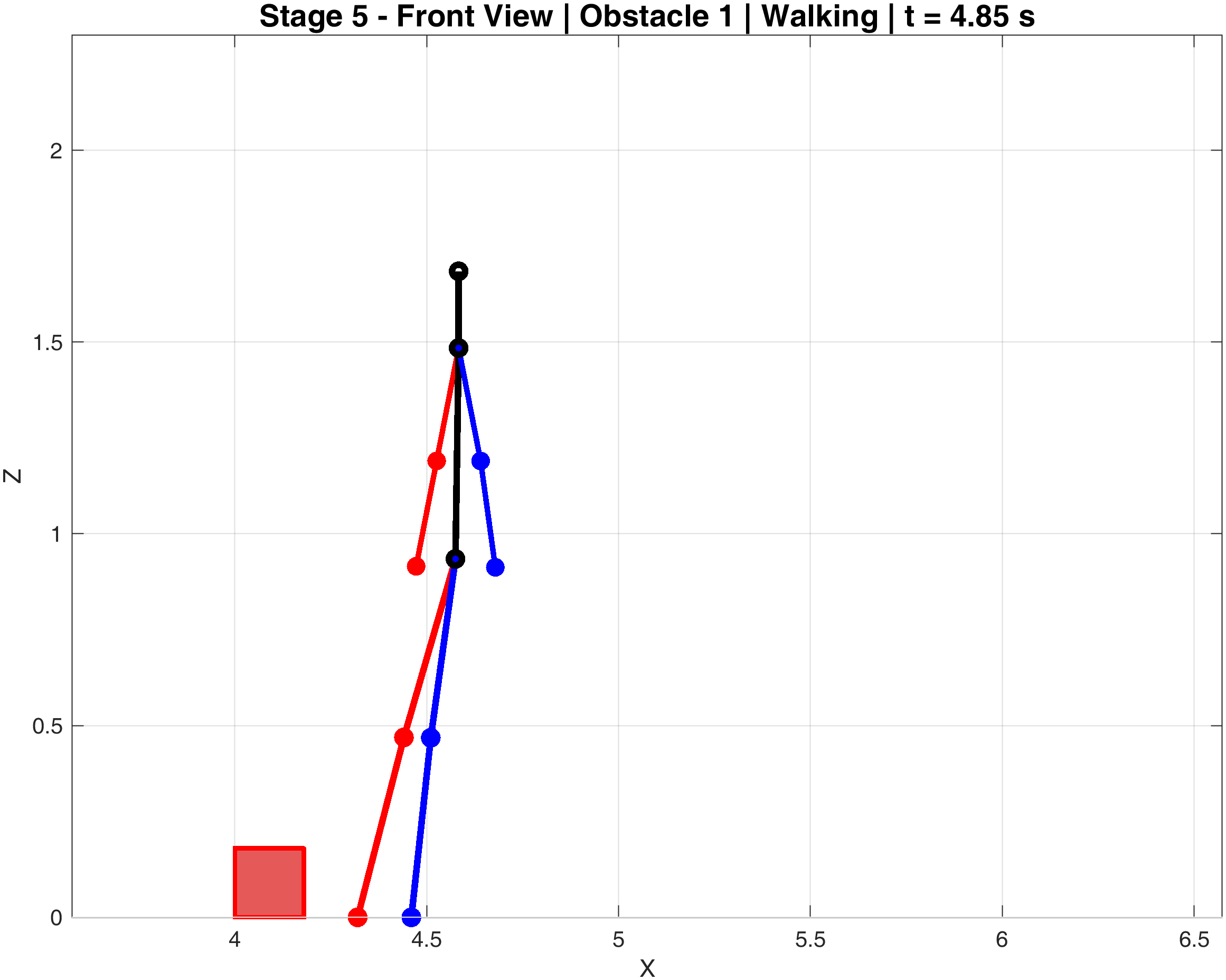}
        \caption{Stage 5}
    \end{subfigure}
    \caption{Front-view crossing sequence for O1 in walking mode,
    showing the five-stage progression with a commanded foot
    apex of \(0.266~\mathrm{m}\).}
    \label{fig:obs1_sequence}
\end{figure*}

\begin{figure*}[!t]
    \centering
    \begin{subfigure}[t]{0.19\textwidth}
        \centering
        \includegraphics[width=\linewidth]{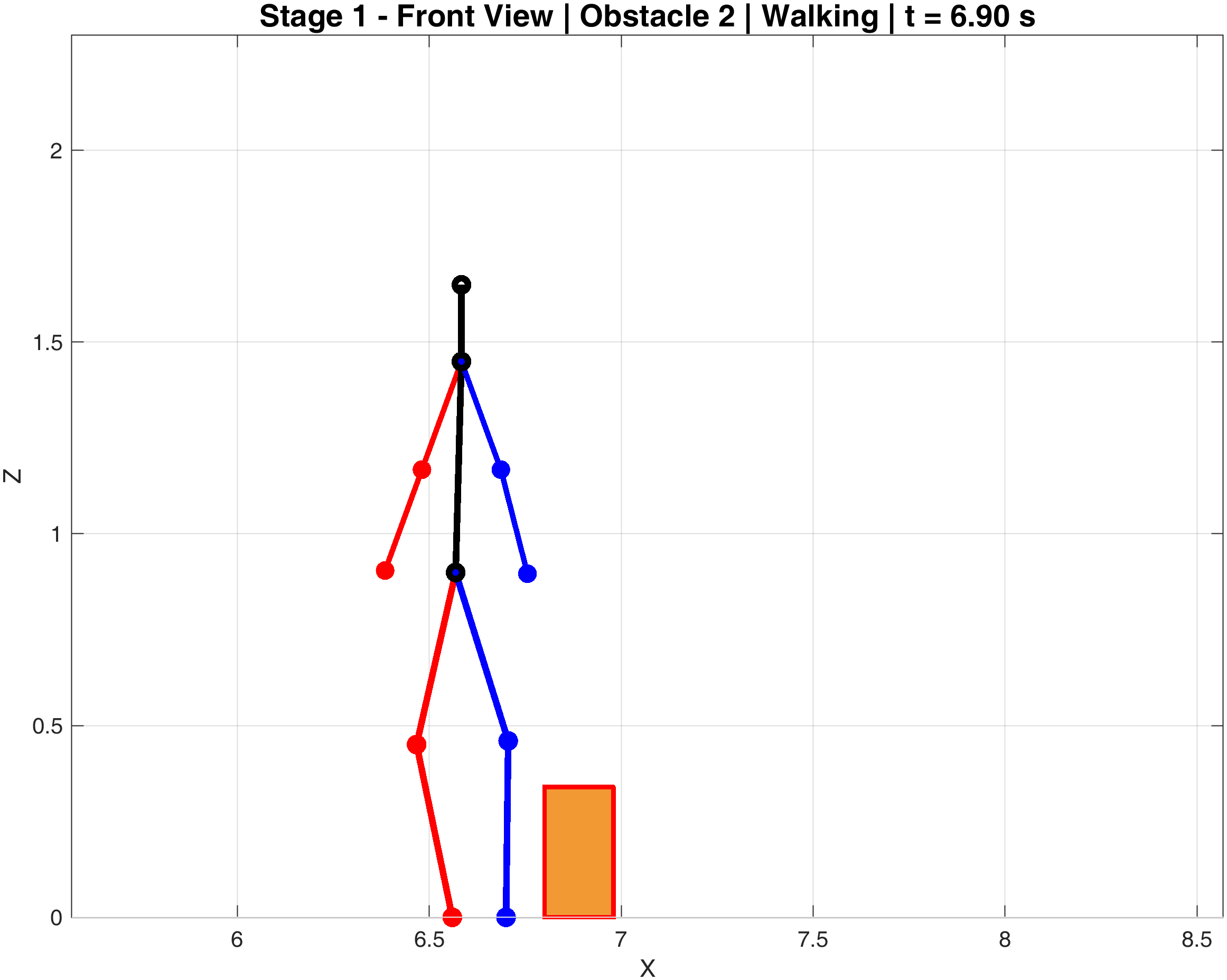}
        \caption{Stage 1}
    \end{subfigure}
    \hfill
    \begin{subfigure}[t]{0.19\textwidth}
        \centering
        \includegraphics[width=\linewidth]{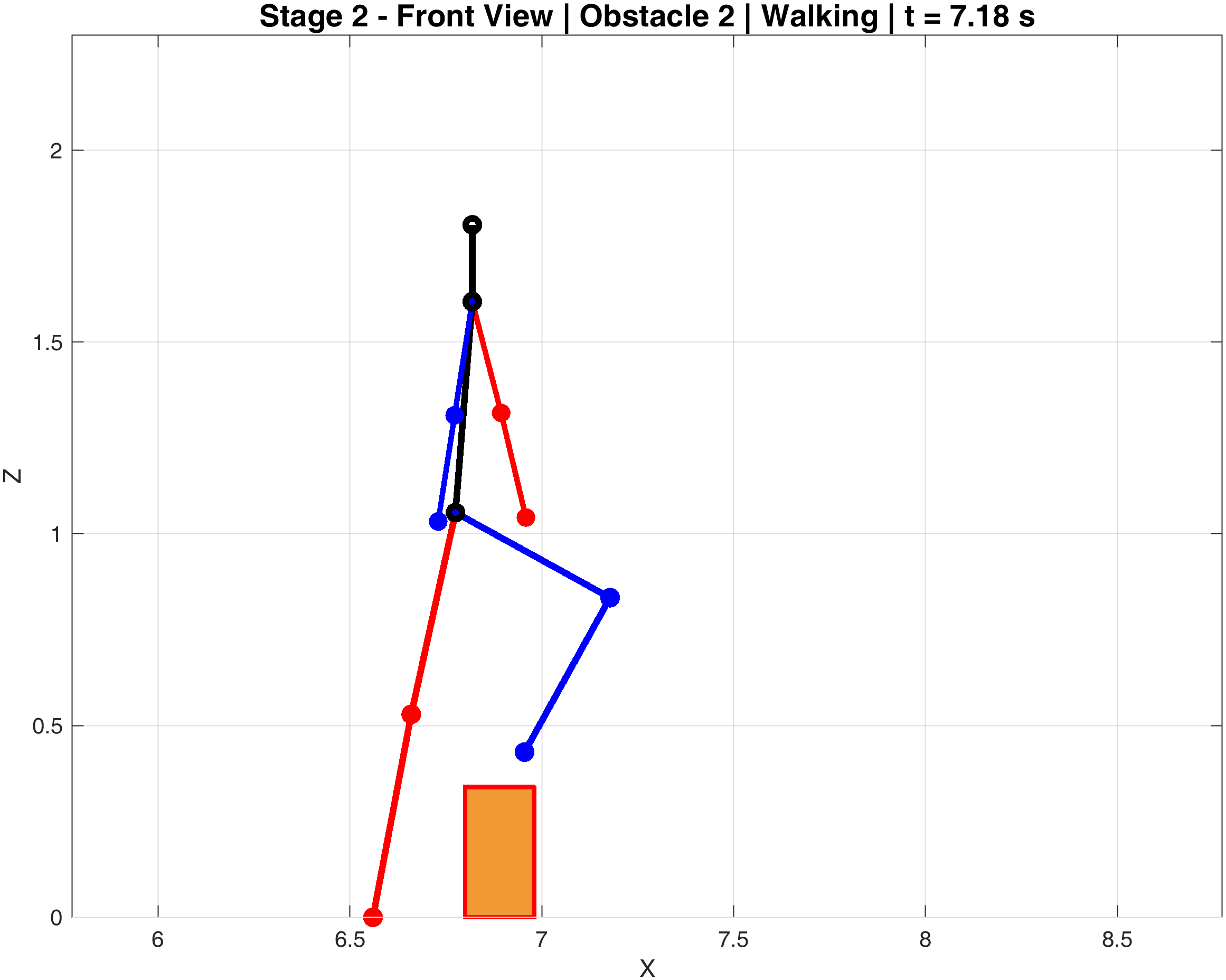}
        \caption{Stage 2}
    \end{subfigure}
    \hfill
    \begin{subfigure}[t]{0.19\textwidth}
        \centering
        \includegraphics[width=\linewidth]{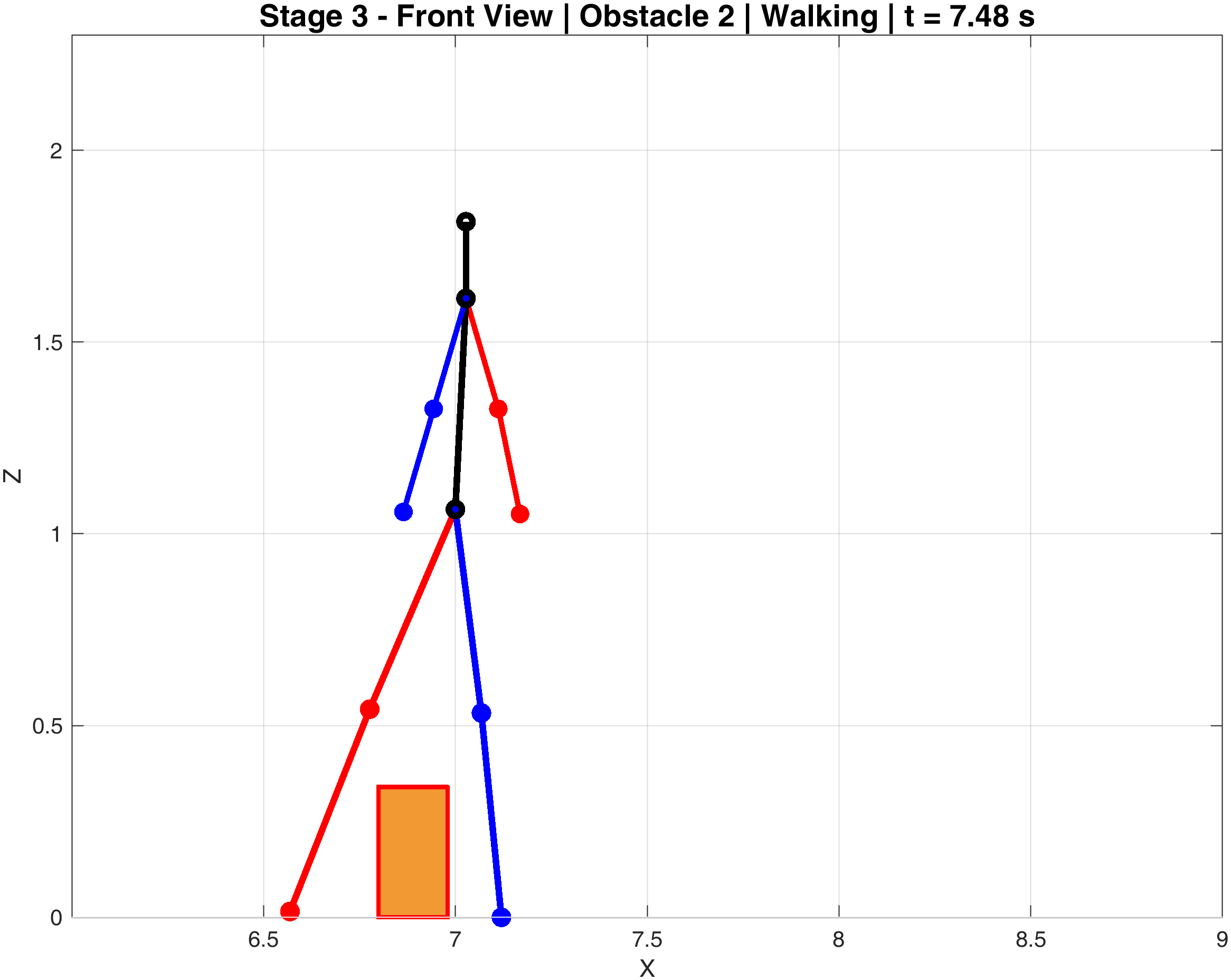}
        \caption{Stage 3}
    \end{subfigure}
    \hfill
    \begin{subfigure}[t]{0.19\textwidth}
        \centering
        \includegraphics[width=\linewidth]{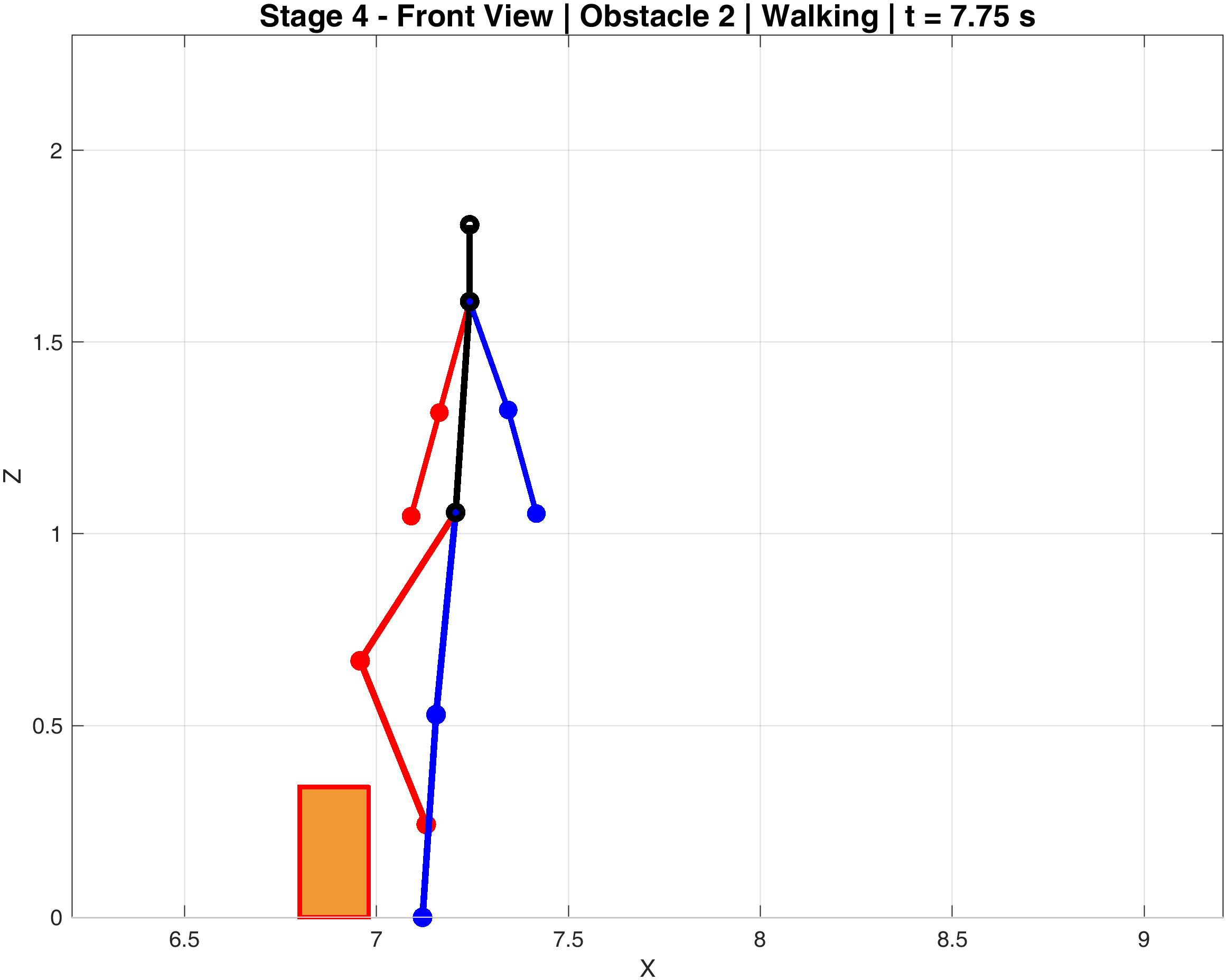}
        \caption{Stage 4}
    \end{subfigure}
    \hfill
    \begin{subfigure}[t]{0.19\textwidth}
        \centering
        \includegraphics[width=\linewidth]{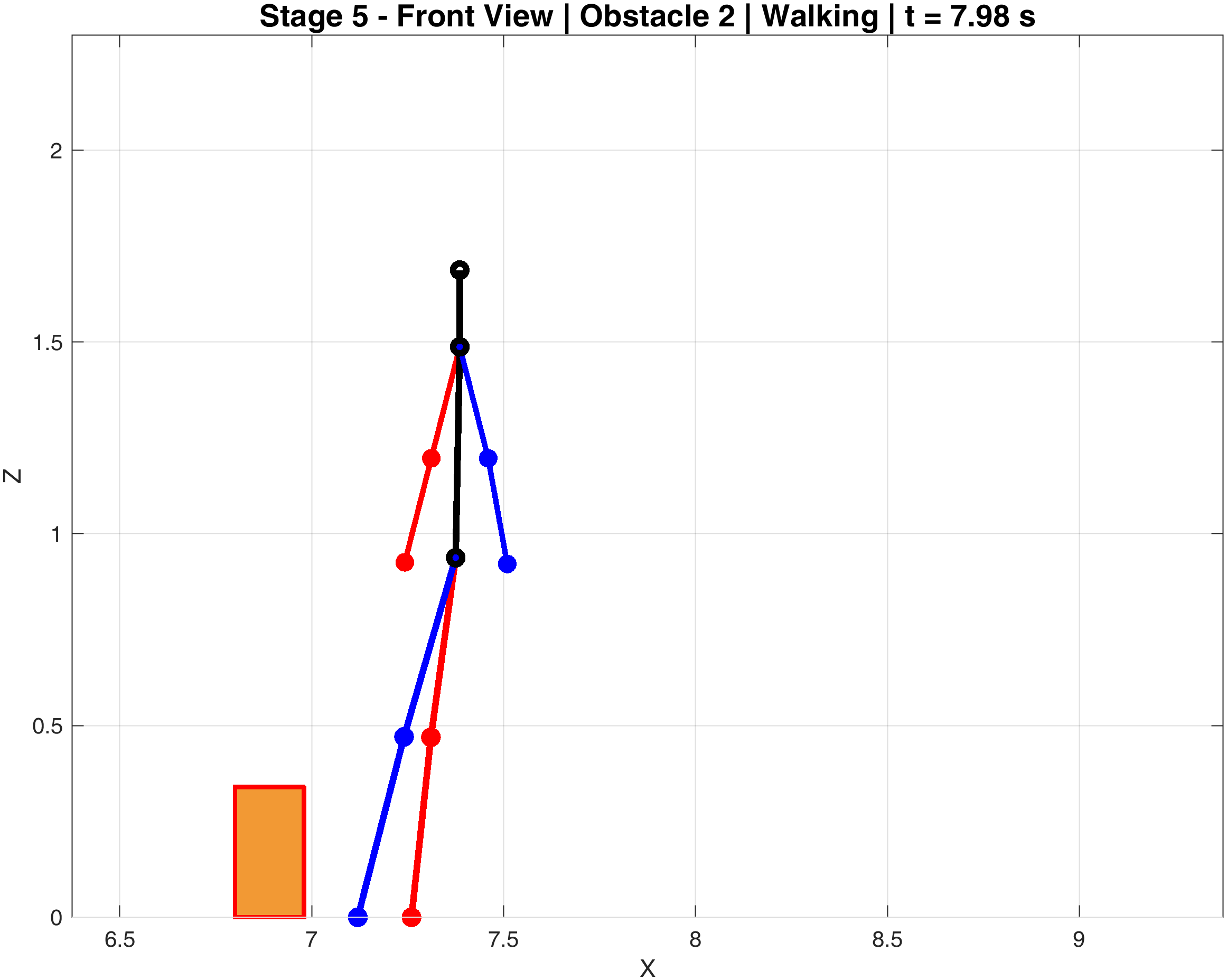}
        \caption{Stage 5}
    \end{subfigure}
    \caption{Front-view crossing sequence for O2 in walking mode.
    The foot apex increases to \(0.458~\mathrm{m}\), while the
    commanded peak half-stance width remains identical to O1.}
    \label{fig:obs2_sequence}
\end{figure*}

\begin{figure*}[!t]
    \centering
    \begin{subfigure}[t]{0.19\textwidth}
        \centering
        \includegraphics[width=\linewidth]{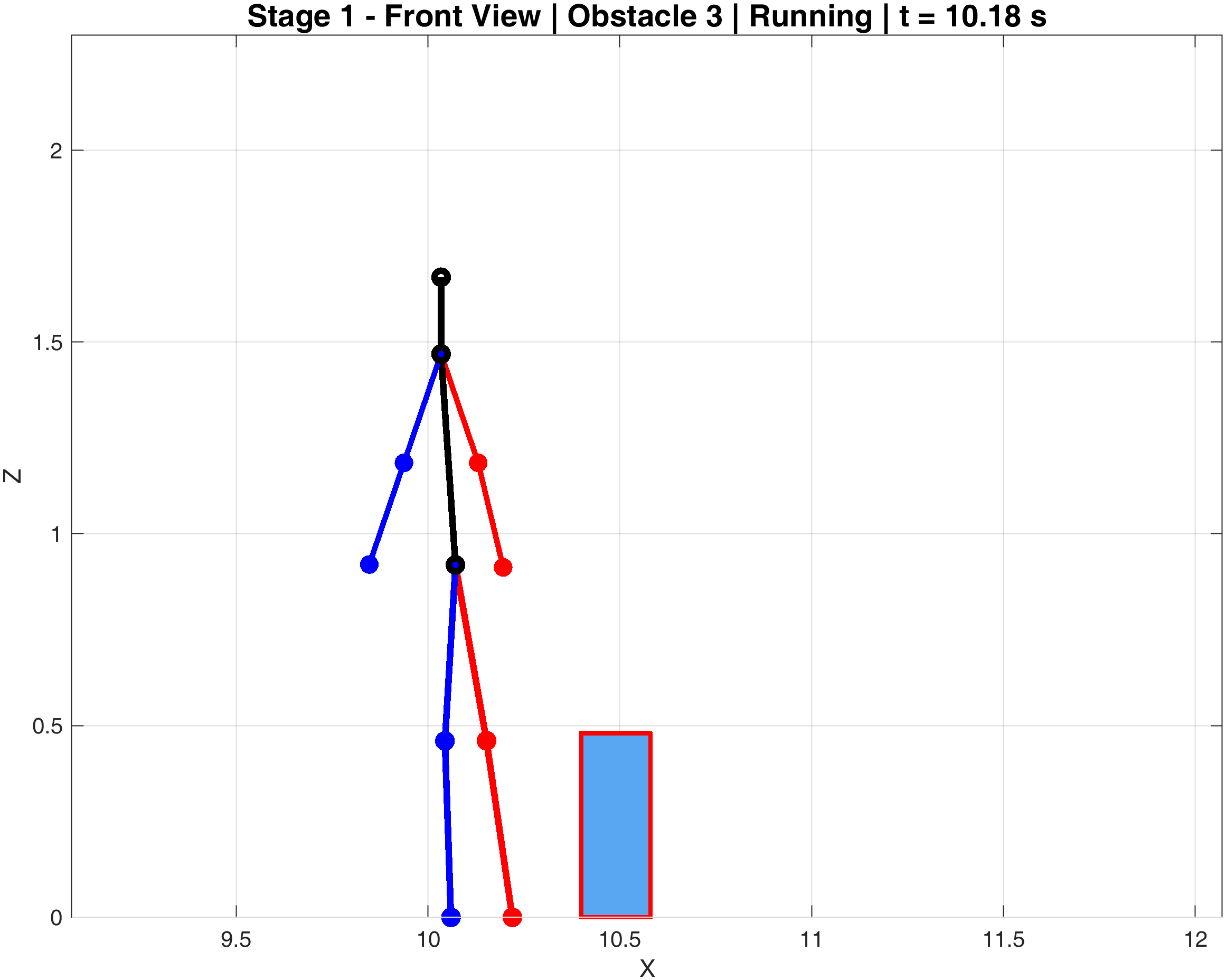}
        \caption{Stage 1}
    \end{subfigure}
    \hfill
    \begin{subfigure}[t]{0.19\textwidth}
        \centering
        \includegraphics[width=\linewidth]{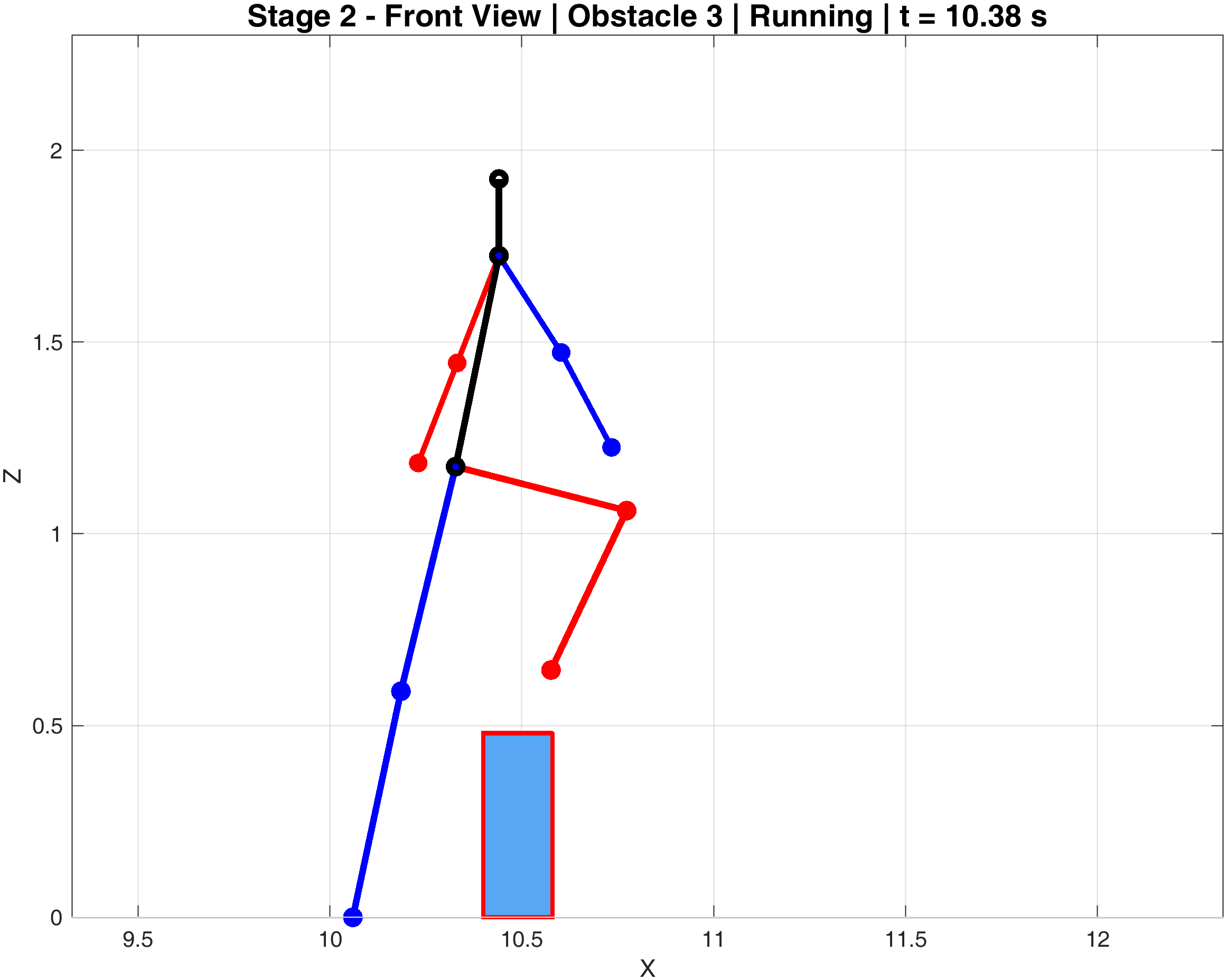}
        \caption{Stage 2}
    \end{subfigure}
    \hfill
    \begin{subfigure}[t]{0.19\textwidth}
        \centering
        \includegraphics[width=\linewidth]{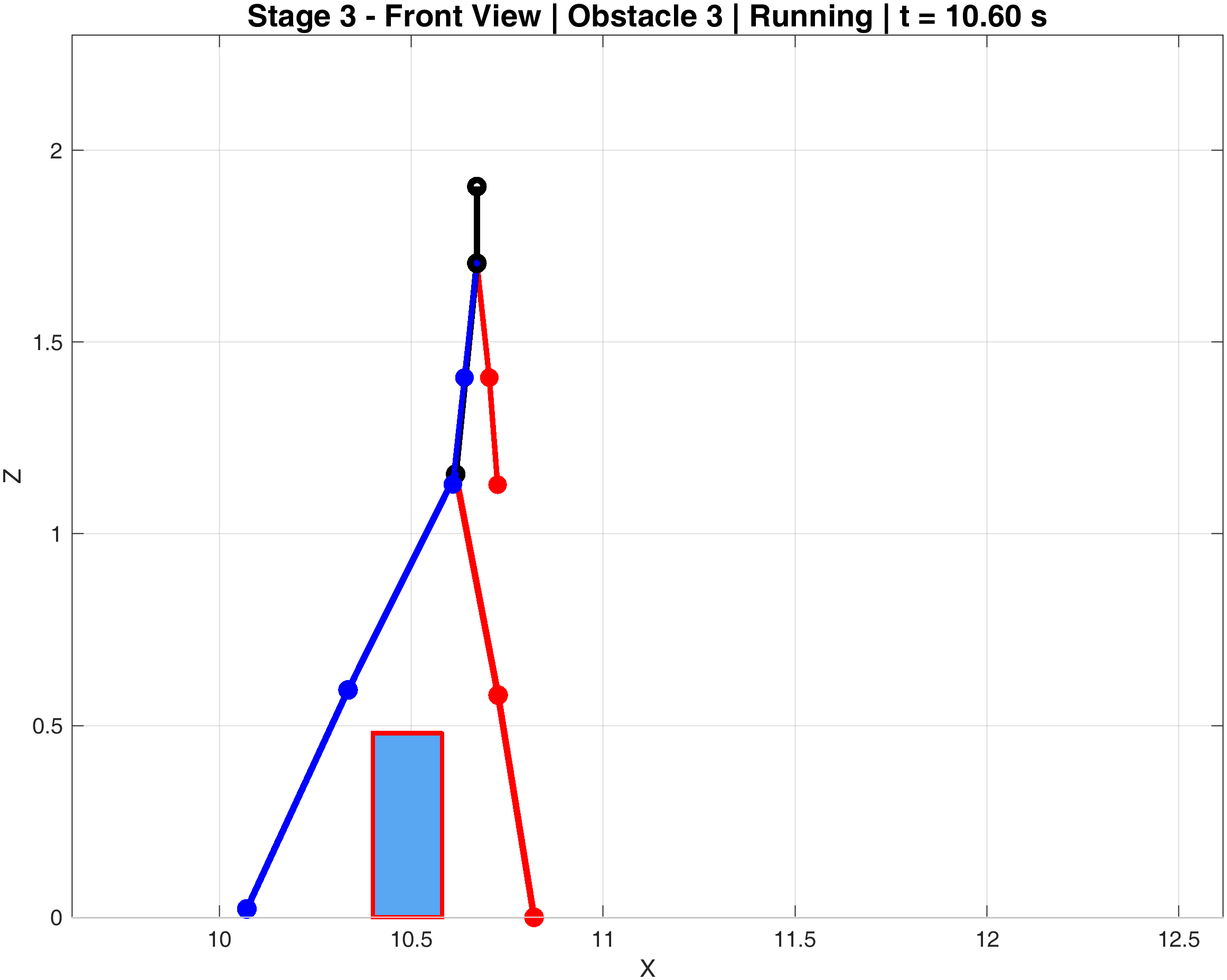}
        \caption{Stage 3}
    \end{subfigure}
    \hfill
    \begin{subfigure}[t]{0.19\textwidth}
        \centering
        \includegraphics[width=\linewidth]{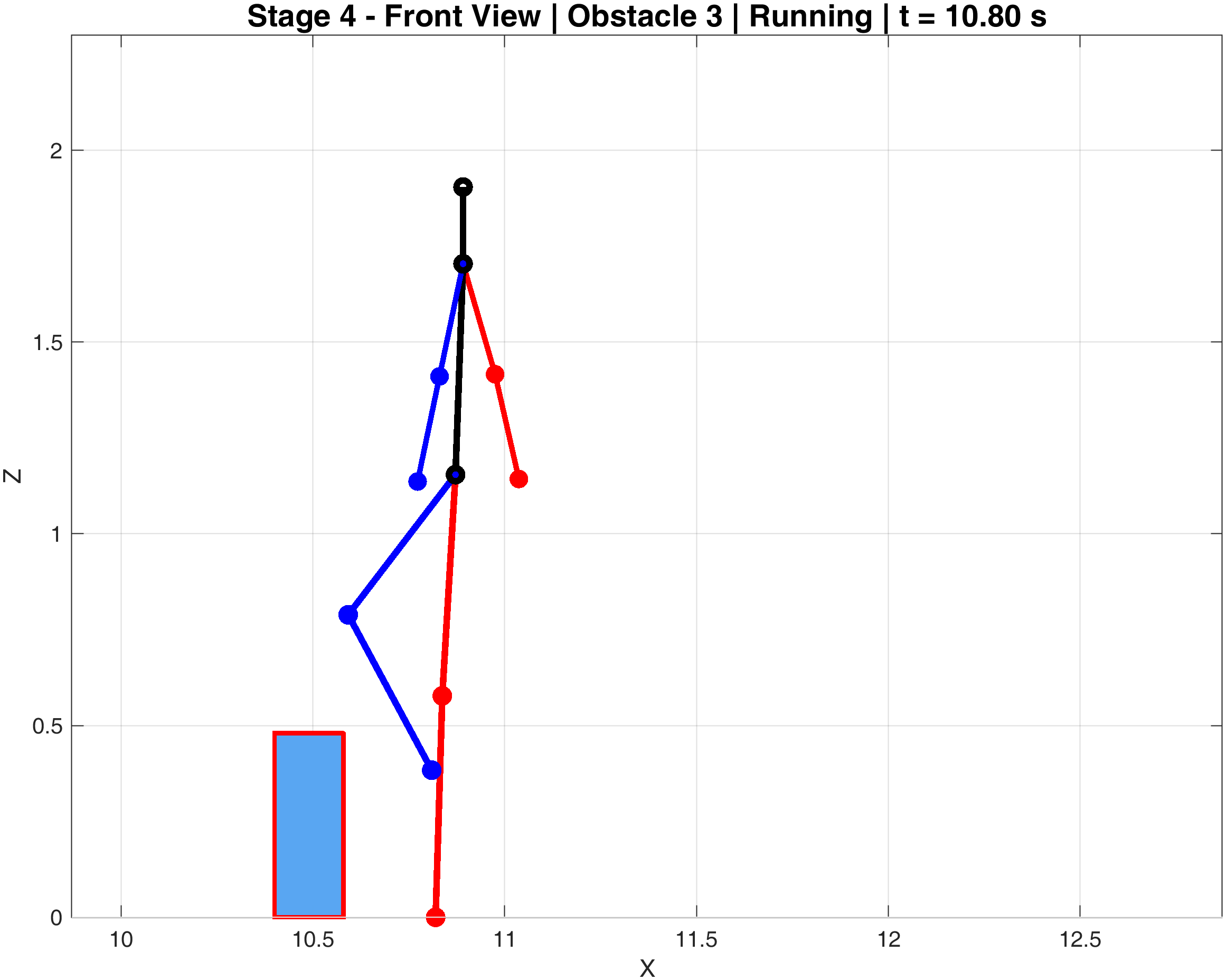}
        \caption{Stage 4}
    \end{subfigure}
    \hfill
    \begin{subfigure}[t]{0.19\textwidth}
        \centering
        \includegraphics[width=\linewidth]{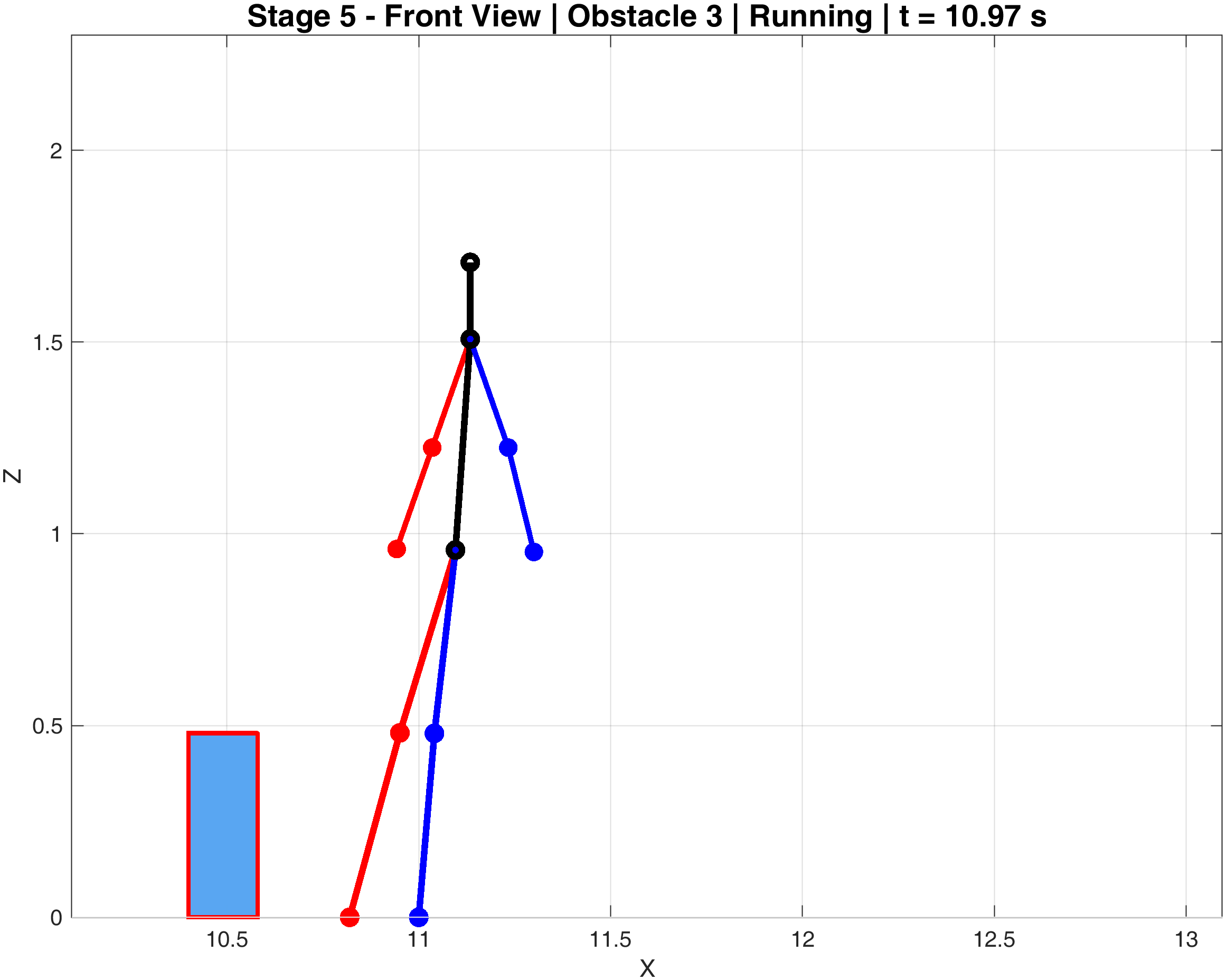}
        \caption{Stage 5}
    \end{subfigure}
    \caption{Front-view crossing sequence for O3 in running mode.
    The ordered crossing stages are maintained at an active
    crossing speed of \(1.28~\mathrm{m/s}\).}
    \label{fig:obs3_sequence}
\end{figure*}

\begin{figure*}[!t]
    \centering
    \begin{subfigure}[t]{0.19\textwidth}
        \centering
        \includegraphics[width=\linewidth]{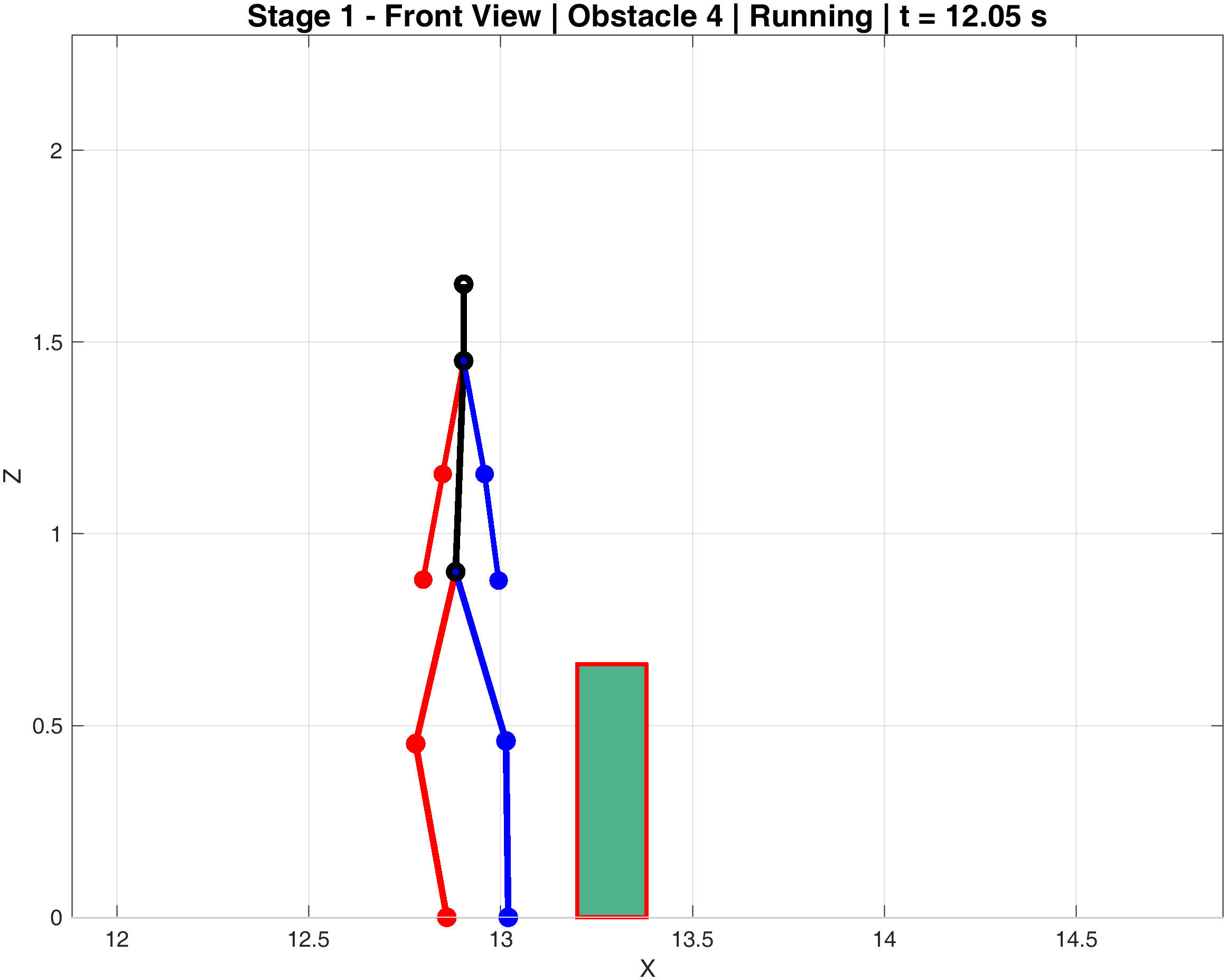}
        \caption{Stage 1}
    \end{subfigure}
    \hfill
    \begin{subfigure}[t]{0.19\textwidth}
        \centering
        \includegraphics[width=\linewidth]{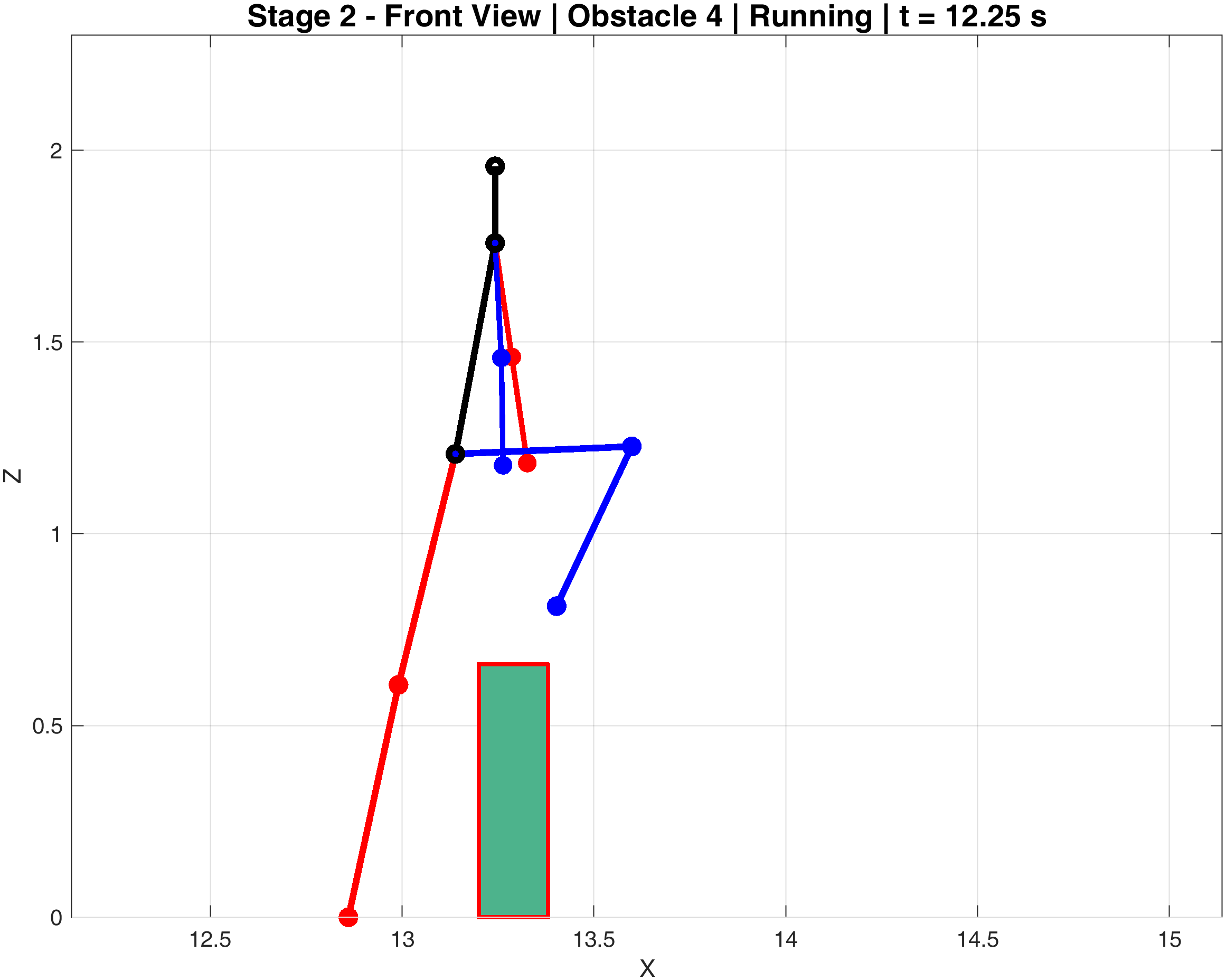}
        \caption{Stage 2}
    \end{subfigure}
    \hfill
    \begin{subfigure}[t]{0.19\textwidth}
        \centering
        \includegraphics[width=\linewidth]{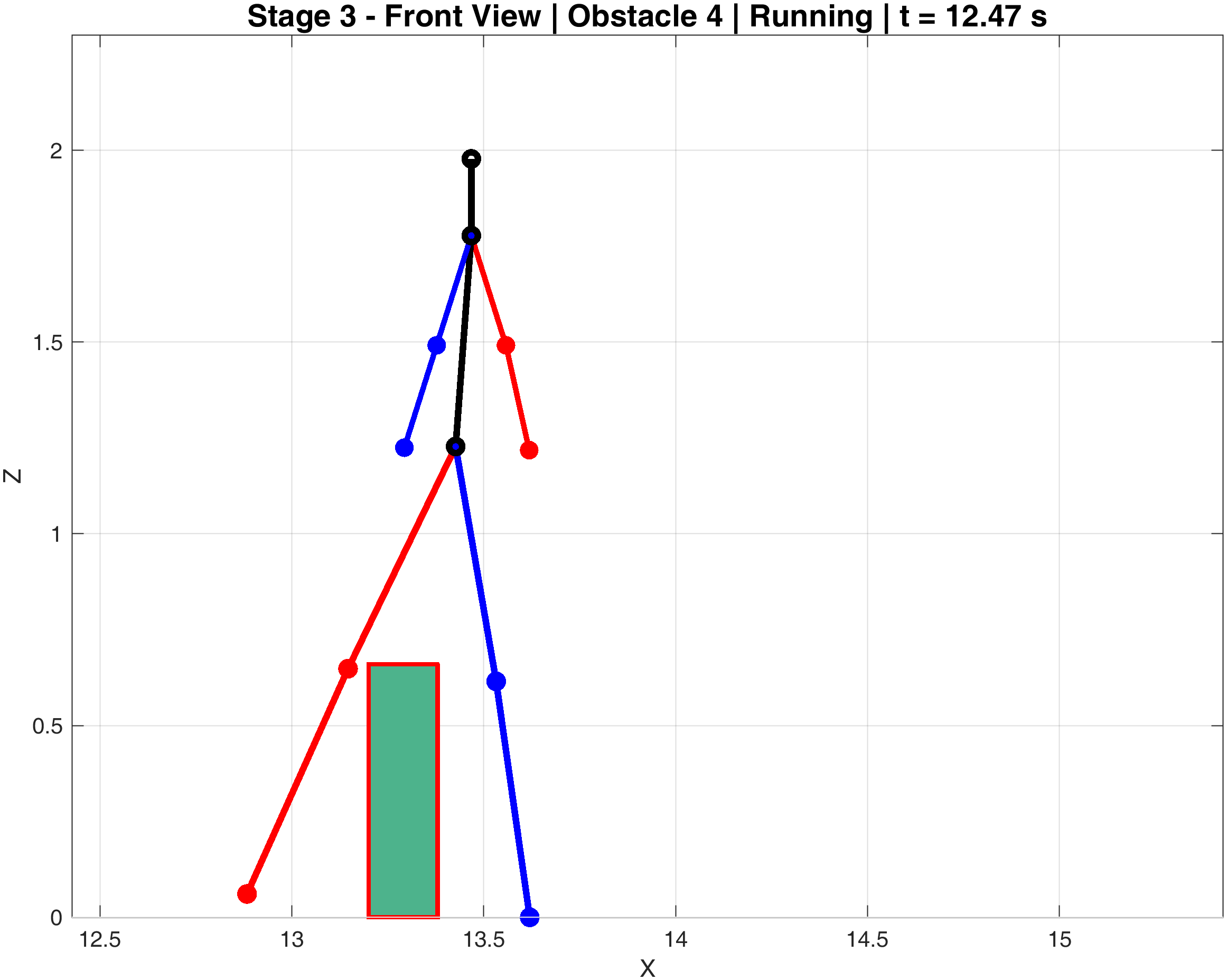}
        \caption{Stage 3}
    \end{subfigure}
    \hfill
    \begin{subfigure}[t]{0.19\textwidth}
        \centering
        \includegraphics[width=\linewidth]{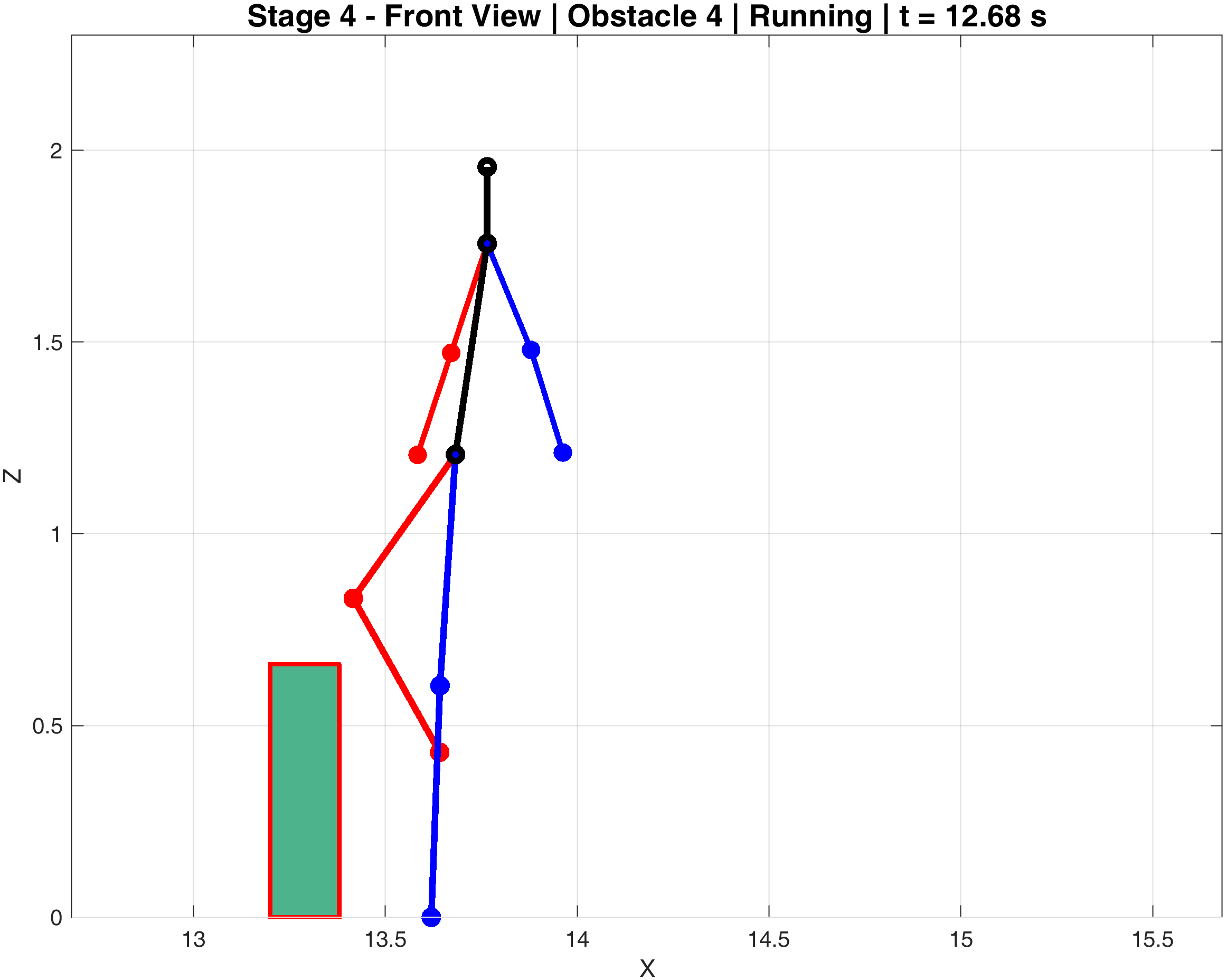}
        \caption{Stage 4}
    \end{subfigure}
    \hfill
    \begin{subfigure}[t]{0.19\textwidth}
        \centering
        \includegraphics[width=\linewidth]{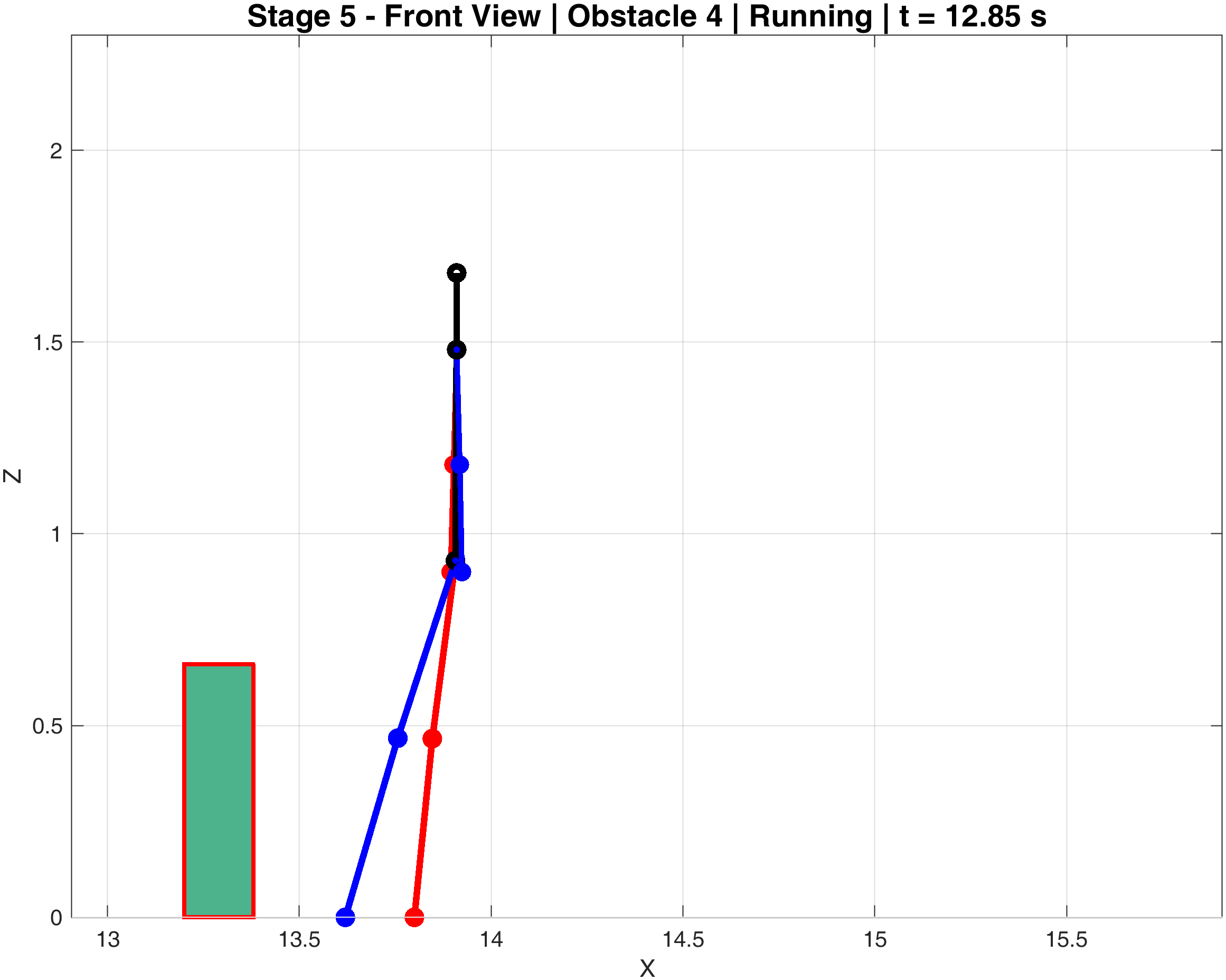}
        \caption{Stage 5}
    \end{subfigure}
    \caption{Front-view crossing sequence for O4 in running mode.
    This is the tallest successfully crossed obstacle in the
    reported sequence, with minimum sampled clearance
    \(0.017~\mathrm{m}\).}
    \label{fig:obs4_sequence}
\end{figure*}

\begin{figure*}[!t]
    \centering
    \begin{subfigure}[t]{0.3\textwidth}
        \centering
        \includegraphics[width=\linewidth]{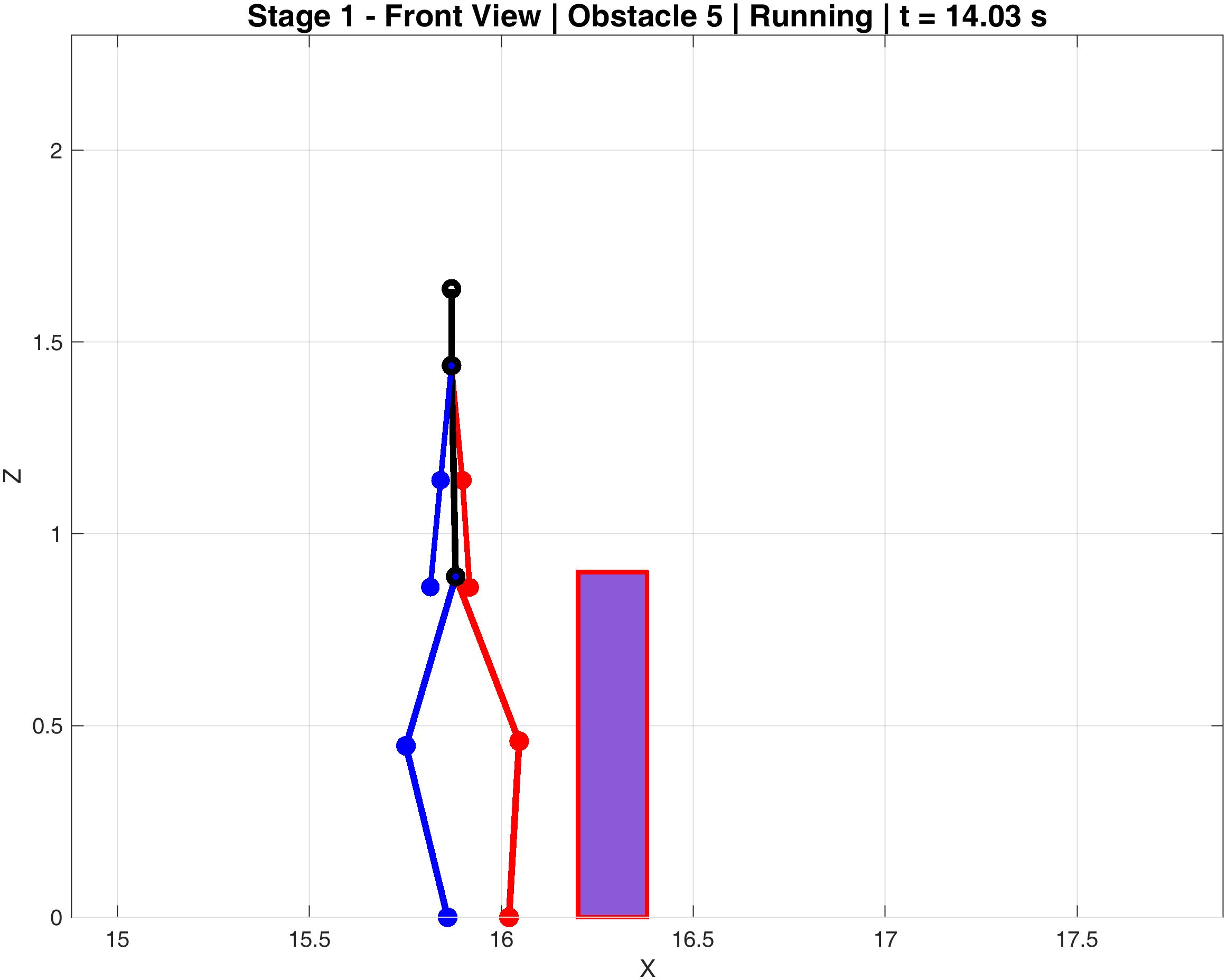}
        \caption{Initial crossing stage.}
    \end{subfigure}
    \hfill
    \begin{subfigure}[t]{0.3\textwidth}
        \centering
        \includegraphics[width=\linewidth]{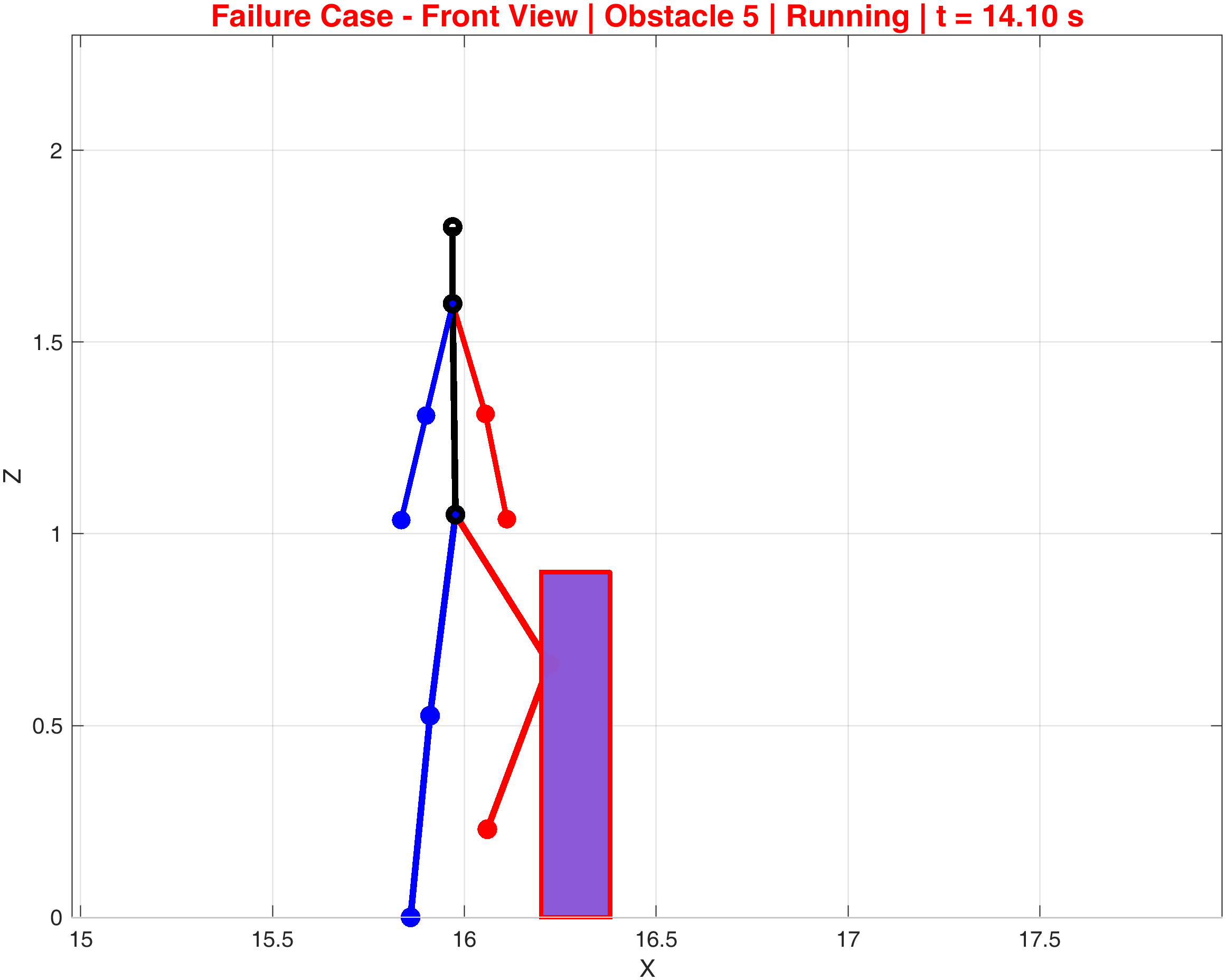}
        \caption{Collision configuration.}
    \end{subfigure}
    \hfill
    \begin{subfigure}[t]{0.3\textwidth}
        \centering
        \includegraphics[width=\linewidth]{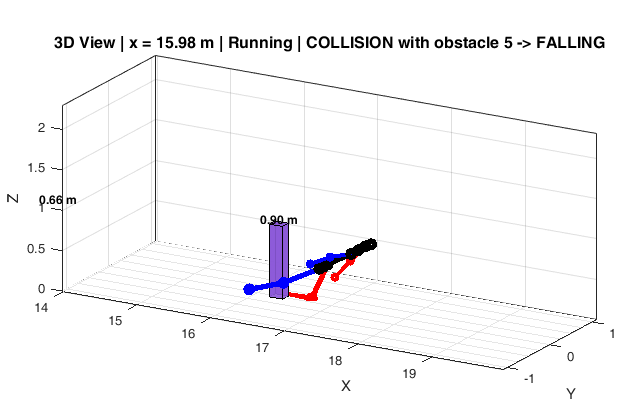}
        \caption{Global 3D view of fall initiation.}
        \label{fig:fall_case_3d}
    \end{subfigure}
    \caption{Attempted O5 crossing in running mode. Collision
    occurs before the intended swing apex is attained; the
    snapshots alone do not isolate the initiating failure
    mechanism.}
    \label{fig:obs5_sequence}
\end{figure*}

The framework was evaluated using a MATLAB-based reduced-order
bipedal locomotion simulation. Five rectangular obstacles were
arranged along the forward direction, with fixed forward extent
\(w_i=0.18~\mathrm{m}\) and transverse extent
\(d_i=0.14~\mathrm{m}\). Obstacle height increased progressively,
as summarized in Table~\ref{tab:sim_setup}. Representative scene
configurations and crossing stages are shown in
Figs.~\ref{fig:obs1_sequence}--\ref{fig:obs5_sequence}.

O1--O2 were encountered in walking mode, whereas O3--O5 followed
the walk-to-run transition. Walking used nominal and active
crossing speeds of \(1.00\) and \(0.75~\mathrm{m/s}\),
respectively; the corresponding running speeds were \(1.75\)
and \(1.28~\mathrm{m/s}\). The MPC sampling interval was
\(0.02~\mathrm{s}\), with a 20-step prediction horizon,
providing \(0.40~\mathrm{s}\) of preview. The same reported
trajectory-generation and MPC parameters were used throughout
the sequence.

\begin{table}[t]
\centering
\caption{Obstacle configuration and commanded crossing references.}
\label{tab:sim_setup}
\resizebox{\columnwidth}{!}{%
\begin{tabular}{c c c c c c c}
\hline
Obs. & Regime & Pos. (m) & Height (m) & Speed (m/s)
& Cmd. apex (m) & Half stance (m) \\
\hline
O1 & Walk & 4.0  & 0.18 & 0.75 & 0.266 & 0.120 \\
O2 & Walk & 6.8  & 0.34 & 0.75 & 0.458 & 0.120 \\
O3 & Run  & 10.4 & 0.48 & 1.28 & 0.675 & 0.120 \\
O4 & Run  & 13.2 & 0.66 & 1.28 & 0.898 & 0.120 \\
O5 & Run  & 16.2 & 0.90 & 1.28 & 1.196 & 0.120 \\
\hline
\end{tabular}%
}
\end{table}

The nominal half-stance width increased from \(0.085\) to
\(0.120~\mathrm{m}\) during crossing, corresponding to
\(41.2\%\) widening. The peak obstacle-induced pelvis offset
was \(0.021~\mathrm{m}\). Because all obstacles had identical
transverse extent, these lateral reference parameters were
unchanged across cases. Consequently, the sequence evaluates
height variation with lateral adaptation enabled, rather than
independently testing depth-dependent adaptation. Height and
gait regime also vary together, preventing isolation of their
individual effects.

Performance was assessed using crossing snapshots, minimum
signed body--obstacle clearance, realized foot apex, maximum
knee flexion, joint trajectories, and collision status. The
five encounters constitute a nominal simulation sequence,
rather than repeated randomized trials.

\section{Results and Discussion}

The evaluation examines whether the phase-structured references
and predictive tracking maintain coordinated traversal as
obstacle height increases across walking and running regimes.
Successful and failed encounters are retained to characterize
the demonstrated operating conditions without implying
performance beyond the tested sequence.

\subsection{Obstacle-Crossing Performance}

Figures~\ref{fig:obs1_sequence}--\ref{fig:obs4_sequence} show
coordinated leading-leg clearance, straddle, trailing-leg
clearance, and recovery for the successful encounters.
O1--O2 use walking references, whereas O3--O4 retain the
crossing sequence under running references. O4 is the most
demanding successful encounter by obstacle height, with a
smaller clearance margin than O1--O3.
Figure~\ref{fig:obs5_sequence} documents the failed O5 attempt.

\begin{figure}[t]
    \centering
    \begin{subfigure}[t]{0.48\linewidth}
        \centering
        \includegraphics[width=\linewidth]{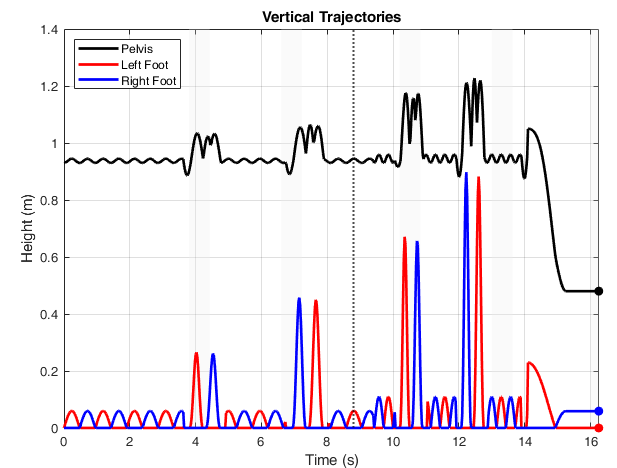}
        \caption{}
        \label{fig:vertical_traj}
    \end{subfigure}
    \hfill
    \begin{subfigure}[t]{0.48\linewidth}
        \centering
        \includegraphics[width=\linewidth]{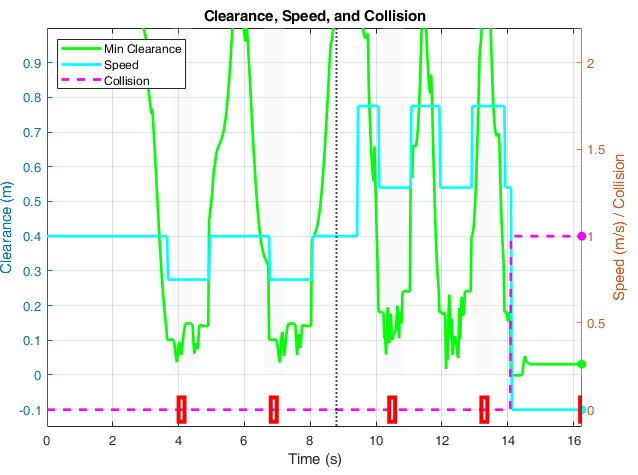}
        \caption{}
        \label{fig:clearance_speed_collision}
    \end{subfigure}
    \caption{Global simulation diagnostics: (a) vertical pelvis
    and foot trajectories; (b) minimum signed obstacle clearance,
    locomotion speed, and collision indicator.}
    \label{fig:traj_clearance_combined}
\end{figure}

Figure~\ref{fig:traj_clearance_combined} shows increasing
realized swing-foot elevation across O1--O4. Their minimum
sampled clearances are \(0.036\), \(0.036\), \(0.059\), and
\(0.017~\mathrm{m}\), respectively. O5 yields
\(-0.001~\mathrm{m}\), indicating penetration under the
adopted collision metric. Clearance does not increase
monotonically with foot apex, illustrating that the closest
body point also depends on configuration and crossing timing.

\begin{figure}[t]
    \centering
    \includegraphics[width=0.95\linewidth]{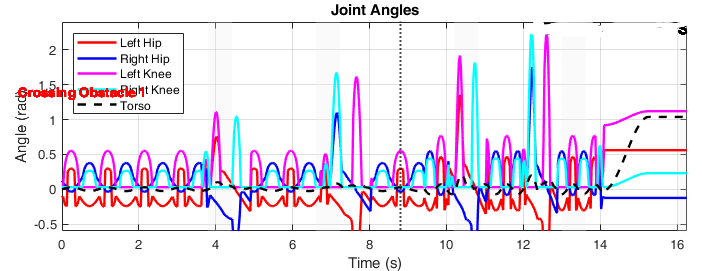}
    \caption{Joint-angle evolution over the locomotion sequence.
    The reported maximum knee flexion increases across the
    successful O1--O4 encounters.}
    \label{fig:joint_angles}
\end{figure}

\subsection{Quantitative Analysis and Observed Failure}

Table~\ref{tab:quant_results} summarizes the recorded outcomes.
The realized foot apex increases from \(0.266~\mathrm{m}\)
for O1 to \(0.898~\mathrm{m}\) for O4, while maximum knee
flexion increases from \(63.5^{\circ}\) to \(130.8^{\circ}\).
These trends are consistent with increasing vertical
clearance demand. O3 attains \(0.671~\mathrm{m}\), compared
with its commanded \(0.675~\mathrm{m}\); the other successful
apex values match their commands at the reported precision.
These peak comparisons do not constitute trajectory-wide
tracking-error measurements.

\begin{table}[t]
\centering
\caption{Reported obstacle-wise crossing outcomes.}
\label{tab:quant_results}
\resizebox{\columnwidth}{!}{%
\begin{tabular}{c c c c c c}
\hline
Obs. & Regime & Min. clr. (m) & Peak foot (m)
& Max knee ($^\circ$) & Outcome \\
\hline
O1 & Walk & 0.036  & 0.266 & 63.5  & Success \\
O2 & Walk & 0.036  & 0.458 & 95.9  & Success \\
O3 & Run  & 0.059  & 0.671 & 110.0 & Success \\
O4 & Run  & 0.017  & 0.898 & 130.8 & Success \\
O5 & Run  & -0.001 & 0.230 & 52.7  & Failure \\
\hline
\end{tabular}%
}
\end{table}

For O5, the recorded foot apex is \(0.230~\mathrm{m}\),
despite a commanded apex of \(1.196~\mathrm{m}\). This is a
collision-truncated attempt, so the difference must not be
interpreted as a completed-cycle apex tracking error.
The evidence establishes collision before the intended
elevated posture develops, but does not distinguish swing
timing, tracking limitations, prediction error, command
bounds, or kinematic reachability as the primary cause.

At \(1.28~\mathrm{m/s}\), the \(0.40~\mathrm{s}\) horizon
corresponds to nominal forward travel of \(0.512~\mathrm{m}\).
This quantifies preview distance without establishing
insufficient preview as the cause of failure. Contact-point
identity, phase at collision, reference/actual trajectories,
input saturation, and solver status are needed for causal
diagnosis. Speed reduction, earlier swing initiation, and
longer preview remain candidate interventions requiring
controlled tests.

\subsection{Scope of Geometry-Aware Adaptation}

The results demonstrate nominal traversal with the combined
reference generator and MPC enabled.
Table~\ref{tab:quant_results} reports obstacle outcomes,
not comparisons between controller variants. Because all
obstacles share the same transverse extent, the sequence
cannot quantify the independent benefit of stance widening
or pelvis repositioning.

Matched evaluations should compare the full framework with
fixed-swing and height-only baselines, followed by separate
removal of height adaptation, stance widening, and pelvis
offset. Replacing MPC with a tuned nonpredictive tracker
would assess the contribution of predictive tracking.
Such tests must use matched initial conditions, command
limits, geometry, and collision evaluation, reporting
completion, clearance, tracking error, and touchdown velocity.
No numerical advantage over these alternatives is established
by the present results.

\subsection{Discussion}

The reported sequence supports the feasibility of coordinated
geometry-aware references for O1--O4 and retains O5 as an
observed failure. The highest successful height,
\(0.66~\mathrm{m}\), and failed height, \(0.90~\mathrm{m}\),
do not locate an exact critical obstacle height because gait
phase and sequence state may influence the outcome.
Likewise, four successful encounters do not establish a
statistical success rate.

Validation remains limited to reduced-order simulation.
Full rigid-body contact dynamics, compliant interaction,
actuator feasibility, perception uncertainty, and hardware
performance are not established. Robustness to speed,
independent geometry variation, initial phase, and controller
parameters also requires dedicated sweeps. These limitations
motivate comparative evaluation and failure diagnostics
before extension to contact-aware whole-body control and
physical-platform validation.

\section{Conclusion}
A five-phase geometry-aware reference generator was coupled to constrained predictive tracking for walking and running obstacle-crossing references. In the reported reduced-order sequence, O1--O4 were crossed with positive minimum sampled clearance, while O5 collided before attaining the intended foot elevation. The central contribution is coordinated geometric reference construction, supported by nominal feasibility and an explicitly retained failure case. Baseline superiority and the separate effectiveness of transverse adaptation are not established.

Future work proceeds in three stages: first, matched baselines/ablations and independent geometry, speed, and phase sweeps; second, O5 diagnostics and full-body simulation including contact, reachability, and actuator limits; third, perception uncertainty and physical-platform validation. Speed scheduling or longer preview will be assessed as candidate remedies after the failure mechanism is isolated.


\end{document}